\documentclass[11pt]{article}

\usepackage[final]{acl}

\usepackage{times}
\usepackage{latexsym}

\usepackage[T1]{fontenc}

\usepackage[utf8]{inputenc}

\usepackage{microtype}

\usepackage{inconsolata}
\usepackage{booktabs}

\usepackage{graphicx}
\usepackage{url}
\usepackage{multirow}
\usepackage{makecell}
\usepackage{makebox}
\usepackage{enumitem}
\usepackage[table,xcdraw]{xcolor}
\newcommand{\ignore}[1]{}
\usepackage{tcolorbox}
\definecolor{gray}{RGB}{240,240,240}
\definecolor{darkred}{RGB}{255,128,128}
\definecolor{lightred}{RGB}{255,217,217}

\usepackage{tabularx}
\usepackage{array}
\usepackage{amsmath}
\usepackage{amssymb}
\usepackage{listings}

\title{ESC-Skills: Discovering and Self-Evolving Skills \\for Emotional Support Conversations}

\author{Jie Zhu$^{1,2}$, Huaixia Dou$^2$, Shuo Jiang$^{2}$, Junhui Li$^1$\thanks{Corresponding Author.}, Lifan Guo$^2$, \\ 
\textbf{Feng Chen$^2$, Chi Zhang$^2$, Fang Kong$^1$}  \\
$^1$School of Computer Science and Technology, Soochow University \\
$^2$Qwen DianJin Team, Alibaba Cloud Computing\\
{zhujie951121@gmail.com}
}

\begin{document}
\maketitle

\begin{abstract}
Existing emotional support conversation (ESC) systems mainly rely on end-to-end response generation or coarse strategy supervision, offering limited interpretability and little support for systematic skill improvement. We propose \textbf{ESC-Skills}, a skill-centric framework that discovers and self-evolves executable emotional support skills. We first model localized support interactions as \textbf{Intervention Units (IUs)}, which capture state--action--outcome dynamics between seeker states, support interventions, and post-response emotional changes. Based on IUs extracted from both successful and failed ESC dialogues, we construct the \textbf{ESC-Skills Bank}, a repository of executable emotional support skills containing intervention guidance, applicability conditions, expected outcomes, and potential risks. To further improve robustness, we introduce a multi-profile self-evolutionary refinement framework in which an ESC agent interacts with diverse simulated seeker profiles under SAGE evaluation. The resulting interaction traces are analyzed to identify missing skills, unsafe interventions, and profile-specific failure patterns, which are then used to refine the Skills Bank through simulation-based verification. Experimental results demonstrate that ESC-Skills improves both response-level quality and dialogue-level emotional outcomes while providing more interpretable and controllable support behaviors. We will release the code, prompts, and ESC-Skills Bank at \url{https://github.com/aliyun/qwen-dianjin}.
\end{abstract}

\section{Introduction}\label{sec:introduction}

\begin{figure}[!t]
\centering
\includegraphics[width=\columnwidth, trim={0cm 0cm 0cm 0cm}]{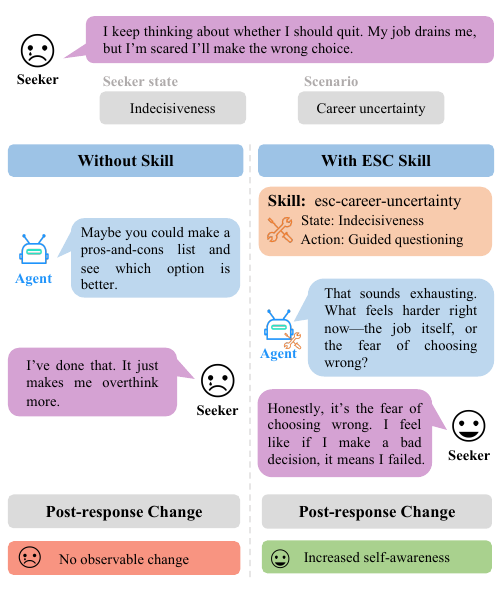}
\caption{Example responses with and without a support skill. Without a suitable support skill (left), the agent produces a generic response that fails to address the seeker's underlying fear, resulting in no observable emotional improvement. With a state- and scenario-aware ESC skill (right), the agent selects a more suitable intervention that facilitates increased self-awareness and constructive emotional reflection.}
\label{fig:example}
\end{figure}

Emotional support conversation (ESC) systems aim to provide timely, scalable, and accessible support for individuals experiencing stress, anxiety, frustration, or emotional distress~\cite{liu-etal-2021-esconv,zhang-etal-2024-escot}. Recent LLM-based ESC advances have primarily focused on improving empathetic response generation and controllable support strategies through synthetic datasets, chain-of-thought reasoning, retrieval mechanisms, and strategy-guided dialogue modeling~\citep{zheng-etal-2023-augesc,zheng-etal-2024-extesc,zhang-etal-2025-intentionesc,ye-etal-2025-sweetiechat,chen-etal-2025-socialsim}. Yet one crucial aspect remains underexplored: how emotional support interventions influence a seeker's subsequent emotional state, and how such intervention knowledge can be explicitly represented, verified, and continually improved over time.

As illustrated in Figure~\ref{fig:example}, although the left case provides a practical suggestion (i.e., \textit{make a pros-and-cons list}) that appears supportive on the surface, it fails to recognize the seeker's underlying self-doubt and fear of failure, resulting in continued rumination and little emotional relief. In contrast, the right case demonstrates how a more appropriate intervention can validate the seeker's emotional burden and guide exploration toward the core source of distress, facilitating constructive post-response changes such as increased self-awareness. These examples suggest that effective ESC depends not only on generating empathetic responses, but also on selecting interventions that induce beneficial emotional state transitions. 

To address this challenge, we propose \textbf{ESC-Skills}, a skill-centric framework for discovering and self-evolving executable emotional support skills. We first formalize localized support interactions as \textbf{Intervention Units (IUs)}, which capture state--action--outcome dynamics between seeker states, support interventions, and post-response emotional changes. Based on IUs extracted from both successful and failed ESC dialogues, we construct the \textbf{ESC-Skills Bank}, a repository of executable emotional support skills containing applicability conditions, intervention guidance, expected outcomes, and potential risk patterns. To further improve skill robustness, we introduce a \textbf{multi-profile self-evolutionary refinement framework} in which an ESC agent interacts with diverse simulated seeker profiles under SAGE evaluation~\cite{zhang-etal-2026-SAGE}. The resulting interaction traces are analyzed to identify missing skills, unsafe interventions, and profile-specific failure patterns, while candidate skill refinements and newly proposed skills are validated through simulation-based verification. Experimental results on ESConv and SAGE show that ESC-Skills improves both response-level quality and long-horizon emotional support outcomes while providing more interpretable and controllable intervention behaviors.

Overall, this paper makes the following contributions:
\begin{itemize}[leftmargin=*]
    \item We propose a skill-centric formulation of ESC based on \textbf{Intervention Units (IUs)}, modeling emotional support as localized state--action--outcome intervention dynamics.
    \item We construct the \textbf{ESC-Skills Bank}, an executable repository of emotional support skills induced from both successful and failed ESC dialogues, capturing effective intervention patterns as well as failure-prone anti-patterns.
    \item We introduce a \textbf{multi-profile self-evolutionary refinement framework} that enables continual skill refinement for ESC agents through simulation-based verification. To the best of our knowledge, this is the first work to develop a self-evolving executable skill framework for ESC. 
\end{itemize}

\section{Related Work}
\paragraph{Emotional support conversations}
Since the release of ESConv~\citep{liu-etal-2021-esconv}, ESC research has largely followed a strategy-predict-then-generate paradigm. Early work improves strategy selection with external commonsense~\citep{tu-etal-2022-misc,cheng-etal-2023-pal}, models turn-level state transitions for global strategy planning~\citep{cheng-etal-2022-improving,zhao-etal-2023-transesc}, or augments training data with synthesized ESC dialogues~\citep{zheng-etal-2023-augesc,zheng-etal-2024-extesc,ye-etal-2025-sweetiechat,zhu-etal-2026-care-esc}. More recent LLM-based approaches explore chain-of-thought reasoning~\citep{zhang-etal-2024-escot} and multi-agent collaboration~\citep{xu-etal-2025-multiagentesc} for more interpretable or coordinated support. In adjacent multi-turn dialogue settings, SEAD~\citep{dai-etal-2026-sead} studies self-evolving training via curriculum-driven user simulation, but focuses on updating model weights for goal-oriented service tasks. Overall, counselling expertise in prior ESC work is still typically embedded in model parameters or fixed prompting schemes, rather than represented as an explicit, editable resource. To our knowledge, framing such expertise as a modular and self-evolving skill bank that transfers across LLM backbones without fine-tuning remains underexplored.

\paragraph{Self-improving agent skills.}
Recent work explores automatic skill refinement for agents via recursive reinforcement learning~\citep{xia-etal-2026-skillrl}, sandboxed optimization~\citep{liu-etal-2026-skillforge}, self-evolutionary verification~\citep{zhang-etal-2026-coevoskills}, reflective memory updates~\citep{zhou-etal-2026-mementoskills}, and lifecycle governance~\citep{liu-etal-2026-skillsvote}. SkillsBench~\citep{li-etal-2026-skillsbench} shows that closed-loop feedback is critical for effective skill improvement.

However, these methods are developed primarily for domains with relatively clear success signals, whereas emotional support conversations lack a reliable deterministic oracle. They also typically model skills as executable code, tool-use procedures, or prompt-level heuristics, while ESC requires \textit{behavioral} intervention knowledge grounded in the seeker's affective state. Our framework therefore represents expertise as structured \texttt{SKILL.md} packages and evaluates it through simulation-based interaction signals.

\begin{figure*}[t]
\centering
\includegraphics[width=\textwidth]{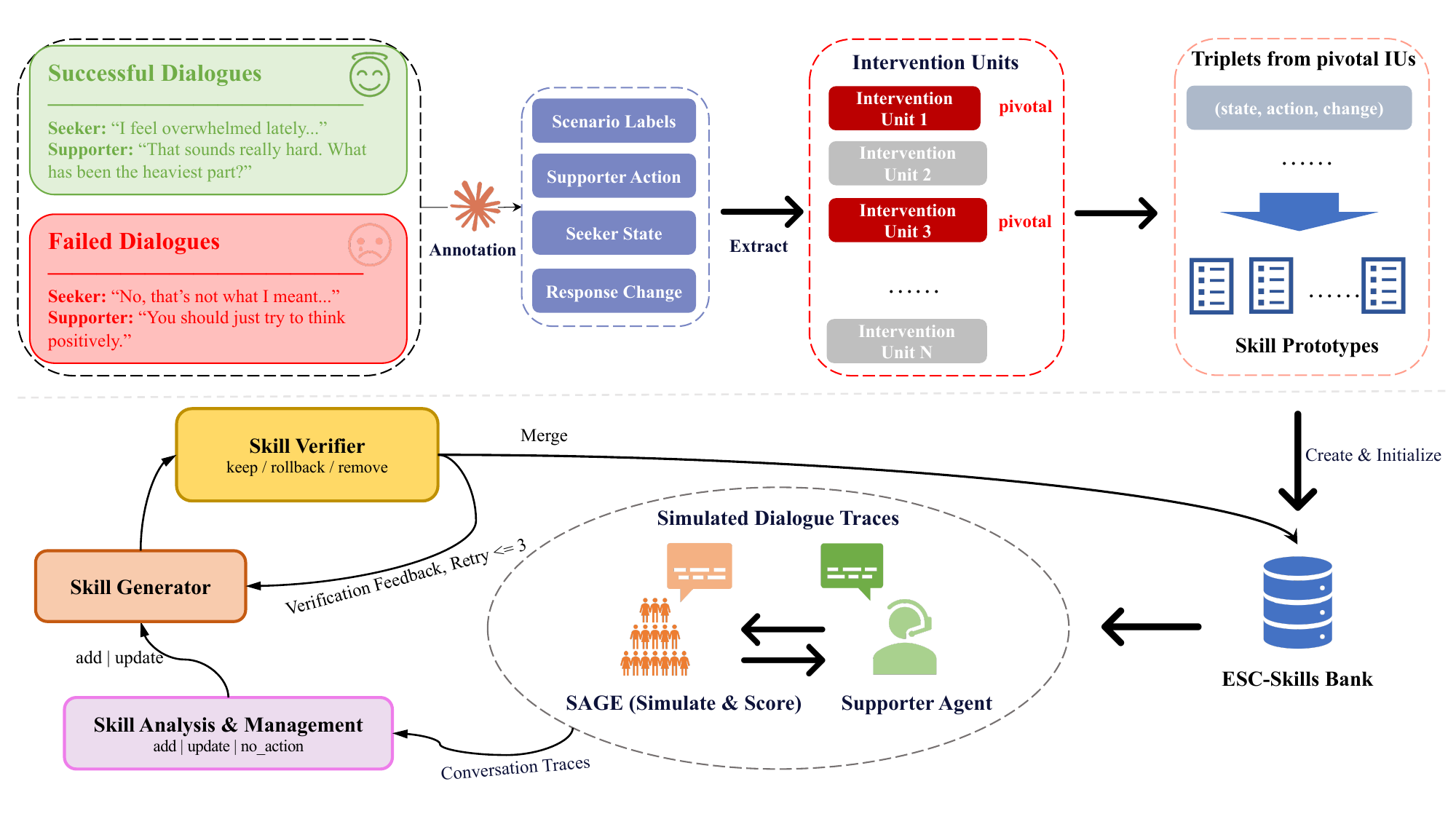}
\caption{Overview of the ESC-Skills Bank construction (upper) and Multi-Profile Self-Evolutionary Skill Refinement framework (lower).}                
\label{fig:overview}
\end{figure*}

\section{Methodology}\label{sec:method}
In this section, we first formalize emotional support conversations as intervention-driven interaction processes and introduce Intervention Units (IUs) for modeling localized state--action--outcome dynamics in Section~\ref{sec:definition}. We then present the construction of the \textbf{ESC-Skills Bank} from annotated intervention patterns in Section~\ref{sec:bank_construction}. Finally, Section~\ref{sec:bank_evolution} introduces a multi-profile self-evolutionary refinement framework that further improves the Skills Bank through interaction-based verification. Figure~\ref{fig:overview} illustrates both the ESC-Skills Bank construction and refinement processes.

\subsection{Problem Definition}\label{sec:definition}
An emotional support conversation (ESC) consists of a multi-turn interaction between a seeker and a supporter, where the supporter aims to provide emotionally appropriate interventions that facilitate constructive emotional changes in the seeker. Formally, let the dialogue context be {\small $U=\{u_1^{usr}, u_1^{sys}, u_2^{usr}, u_2^{sys}, \ldots, u_t^{usr}\}$}, where $u_i^{usr}$ and $u_i^{sys}$ denote the seeker and supporter utterances at turn $i$, respectively. Given the dialogue history $U$ and the current seeker utterance $u_t^{usr}$, the ESC agent generates a supportive response $u_t^{sys}$.

Unlike conventional dialogue generation settings that mainly emphasize response fluency or relevance, we formulate ESC as an intervention-driven process in which response quality is determined by its emotional effect on the seeker. Specifically, we assume that each seeker utterance reflects an underlying emotional state (e.g., self-doubt or emotional distress), while each supporter response corresponds to a support intervention action (e.g., emotional validation or reflective questioning). After the intervention, the seeker transitions to a new emotional state that reflects the post-response emotional effect.

Based on this formulation, we define a localized support interaction as an \textbf{Intervention Unit (IU)}:
\begin{equation}
IU_t = (s_t, a_t, s_{t+1}),
\end{equation}
where $s_t$ denotes the seeker's emotional state before the intervention, $a_t$ denotes the applied support action, and $s_{t+1}$ denotes the resulting emotional state after the intervention. The resulting state transition may reflect either constructive changes (e.g., emotional relief or increased openness) or negative effects (e.g., withdrawal or increased distress).
\ignore{
Given a collection of successful and failed ESC dialogues $\mathcal{D}$, our first objective is to discover reusable emotional support skills from extracted IUs. Each skill represents a generalized state--action--outcome intervention pattern:
\begin{equation}
\kappa = (\mathcal{S}_\kappa, \mathcal{A}_\kappa, \mathcal{O}_\kappa),
\end{equation}
where $\mathcal{S}_\kappa$ denotes the applicability conditions of the skill, $\mathcal{A}_\kappa$ denotes the recommended intervention behavior, and $\mathcal{O}_\kappa$ denotes the expected emotional outcomes and associated risk patterns. These skills are organized into the \textbf{ESC-Skills Bank}, a repository of executable emotional support skills.

However, static skill induction alone is insufficient because emotional support effectiveness varies substantially across different seeker profiles and conversational situations. Therefore, our second objective is to iteratively refine the ESC-Skills Bank through interaction-based verification. Specifically, an ESC agent equipped with the current Skills Bank interacts with diverse simulated seeker profiles, producing dialogue traces that reveal missing capabilities, inappropriate interventions, and profile-specific failure patterns. Based on these traces, the framework performs skill-level updates including skill refinement, addition, revision, and removal. The overall goal of the proposed framework is to continuously improve the quality, robustness, and safety of emotional support interventions through skill discovery and self-evolutionary verification.
}

\subsection{ESC-Skills Bank Construction}\label{sec:bank_construction}
\paragraph{Intervention Unit Extraction.} We use the training split of ESConv (910 conversations) as examples of successful emotional support conversations, and additionally incorporate FailedESConv (196 conversations) as examples of unsuccessful support interactions.\footnote{\url{https://github.com/thu-coai/Emotional-Support-Conversation}} To model intervention dynamics in both successful and failed conversations, we perform multi-dimensional annotation at both the dialogue and utterance levels, including:
\begin{itemize}[leftmargin=*]
\item \textbf{Dialogue-level Scenario Labels.} Each dialogue is assigned one or more scenario labels describing the seeker's real-world situation, such as \textit{loneliness}, \textit{loss and grief}, or \textit{family conflict}. In total, we define 18 scenario categories.
\item \textbf{Utterance-level Seeker States.} Each seeker utterance is annotated with a fine-grained emotional state label, such as \textit{self-blame}, \textit{self-awareness}, or \textit{hopelessness}. In total, we define 15 seeker states.
\item \textbf{Utterance-level Support Actions.} Each supporter response is annotated with an intervention action label describing the underlying support behavior. Compared with the original eight ESConv support strategies, our taxonomy contains 17 types of actions and provides more fine-grained intervention-oriented action descriptions.
\item \textbf{Utterance-level Seeker Response Changes.} For each supporter response, we compare the seeker's emotional states before and after the intervention to identify the resulting post-response emotional change, such as \textit{increased confusion}, \textit{emotional relief}, or \textit{topic shift}.
\end{itemize}

We prompt Claude-Opus to produce these annotations, from which we construct \textbf{Intervention Units (IUs)} for modeling localized state--action--outcome dynamics in ESC. Appendix~\ref{apx:iu_annotation} provides more annotation details.

Based on the annotated response changes, we further categorize IUs into \textit{key IUs} and \textit{non-key IUs}. Key IUs correspond to salient positive or negative emotional shifts in the seeker's post-intervention state, such as \textit{emotional relief}, \textit{more specific expression}, \textit{increased emotional agitation}, or \textit{increased withdrawal}. In contrast, IUs associated with weak or stable changes (e.g., \textit{no observable change}) are treated as non-key IUs. In total, we extract 17,858 IUs, including 10,181 key IUs consisting of 9,697 positive and 484 negative instances. Table~\ref{tbl:iu_structure} in Appendix~\ref{apx:iu_annotation} illustrates the structure of an IU.

\paragraph{Skill Prototype Generation.}
We induce initial emotional support skill prototypes from the extracted key IUs. Specifically, we group key IUs by their \textit{(seeker state, support action)} tuples, where each group captures a recurring intervention pattern under similar emotional conditions. To improve reliability, groups containing fewer than five IUs are discarded. After filtering, we obtain 258 skill prototype groups, each representing a candidate emotional support intervention pattern derived from recurring state--action interactions. Appendix~\ref{apx:prototype} presents examples of skill prototypes.

\paragraph{Skill Bank Construction.} 
The extracted prototypes capture recurring \textit{(seeker state, support action)} intervention patterns, but remain aggregated interaction patterns rather than executable support knowledge. To make them operationally usable, we transform the prototypes into structured emotional support skills and organize them into the \textbf{ESC-Skills Bank}.

First, we cluster the 258 prototypes according to the semantic similarity of their seeker states and support actions, producing recurring emotional support scenarios such as \textit{resistance handling}, \textit{grief and loss}, and \textit{risk awareness}. Each cluster contains related prototypes together with their associated key IUs, preserving both effective and risky intervention patterns. 

Second, for each cluster, we prompt Claude-Opus to synthesize a unified emotional support skill based on:
(i) clustered prototypes with effectiveness statistics and response-change distributions,
(ii) representative dialogue snippets sampled from associated IUs,
and (iii) a predefined skill schema template. Each generated skill is represented as an executable markdown document (\texttt{SKILL.md}) containing structured fields including skill overview, activation conditions, recommended actions, pitfalls to avoid, and representative examples.

Each skill is generated independently using only information from its corresponding cluster, reducing interference across unrelated intervention scenarios. Through this process, we obtain an initial \textbf{ESC-Skills Bank} containing 27 executable emotional support skills, denoted as $\mathcal{B}^{0}$. Appendix~\ref{apx:skill} presents an example skill document.

\subsection{Multi-Profile Self-Evolutionary Skill Refinement}\label{sec:bank_evolution}
Although the initial ESC-Skills Bank $\mathcal{B}^{0}$ captures recurring intervention patterns from ESC dialogues, it is still limited by the coverage and distribution of the training data. Since emotional support effectiveness varies across seeker characteristics and conversational situations, skills induced from static corpora may contain incomplete guidance or hidden failure patterns. To improve robustness and adaptability, we further refine the Skills Bank through a multi-profile interaction framework.

\paragraph{Conversation Simulation.} We use the 500 seeker profiles from RLVER\footnote{\url{https://github.com/Tencent/digitalhuman/tree/main/RLVER}}~\cite{wang-etal-2026-RLVER} and conduct multi-turn ESC simulations under the SAGE framework, where each simulated seeker is initialized with a corresponding profile. During interaction, the ESC agent dynamically retrieves relevant skills from the current Skills Bank according to the seeker's emotional state and dialogue context. Besides the dialogue content, we additionally record turn-level signals including:
(i) the seeker's emotion score and emotional state,
(ii) the scorer's emotional analysis of the agent's response,
and (iii) the seeker's internal thoughts before replying.
These signals provide fine-grained evidence for subsequent analysis. In total, we obtain 500 simulated conversations.

\paragraph{Interaction Analysis.} For each simulated conversation, we prompt Claude-Opus to analyze the applied skills together with their emotional effects on the seeker. The analyzer determines whether the interventions facilitate constructive emotional transitions or instead lead to problematic outcomes such as withdrawal, agitation, confusion, or invalidation. It further identifies whether existing skills require refinement or whether additional skills are needed to address uncovered interaction patterns. Each recommendation is accompanied by explanations grounded in the observed dialogue behaviors and emotional outcomes.

Based on the resulting reports, we aggregate refinement recommendations for existing skills and collect candidate new skills. Similar recommendations are consolidated by Claude-Opus to merge semantically overlapping update reasons and cluster near-duplicate skill proposals. As a result, 9 existing skills are selected for refinement and 12 new skills are identified.

\paragraph{Skill Generation and Verification.} To ensure skill reliability, we introduce a generation--verification refinement loop for both updated and newly proposed skills. For each skill selected for refinement, we prompt Claude-Opus as the \textit{Skill Generator} to produce an updated version conditioned on:
(i) the original \texttt{SKILL.md},
(ii) up to two simulated conversations where the skill leads to problematic outcomes,
and (iii) the lowest-scoring seeker profiles together with their corresponding analysis reports.

For each candidate new skill, we instead provide:
(i) a predefined skill template,
(ii) up to two representative conversations where the new skill is recommended,
and (iii) the associated analysis reports.
The generator then synthesizes a new executable skill following the same schema used in the ESC-Skills Bank.

Let $s$ denote either a refined skill or a newly generated skill. After generation, $s$ is evaluated through simulated interactions using 15 challenging seeker profiles: the lowest-scoring profiles for the original skill, or the globally lowest-scoring profiles for newly added skills. The resulting conversations are evaluated using SAGE. A skill is accepted if either: (i) all verification conversations reach a \textit{Success} state, or (ii) within at most three attempts, its best version achieves a strict improvement in average emotion score. Otherwise, the update is discarded: refined skills are rolled back, while newly proposed skills are removed. The resulting refined ESC-Skills Bank is denoted as $\mathcal{B}^{\star}$, which finally contains 34 emotional support skills. Appendix~\ref{apx:bank-composition} lists the skills in both $\mathcal{B}^{0}$ and $\mathcal{B}^{\star}$.


\section{Experimentation}\label{sec:experimentation}
\subsection{Experimental Settings}\label{sec:setting}
\paragraph{Dataset.} We evaluate ESC-Skills from both response-level and dialogue-level. For response-level evaluation, we use the ESConv dataset~\cite{liu-etal-2021-esconv} by evaluating on the official test split containing 195 emotional support conversations. In this setting, ESC agents generate supportive responses given the dialogue history. This evaluation mainly measures alignment with human supportive behaviors in terms of strategy selection and response quality.

For dialogue-level evaluation, we follow SAGE~\cite{zhang-etal-2026-SAGE} and use its 100 predefined seeker profiles to initialize simulated seekers in multi-turn ESC interactions. Unlike response-level evaluation, SAGE assesses whether ESC agents can sustain constructive long-term emotional support behaviors in extended conversations.

\paragraph{Agent Harness.} We use \textsc{DeerFlow}\footnote{\url{https://github.com/bytedance/deer-flow}}~\cite{bytedance-2026-deerflow}, an open-source long-horizon SuperAgent harness built on LangGraph, as the runtime environment. In experiments, we mainly use its skill-loading mechanism, which loads markdown-format skill files (\texttt{SKILL.md}) from a configurable directory, enabling fair comparison across different skill banks. 

\paragraph{Models.} We evaluate ESC-Skills using multiple LLM backbones, including Qwen3.6-Plus, GPT-5.4-0305-Global, Gemini-3.1-Flash, Claude-Opus-4.6, Claude-Sonnet-4.6, and Claude-Haiku-4.5. 

\paragraph{Baselines.} Besides the \textbf{No-Skill} baseline, where no external skills are provided, we compare ESC-Skills with four representative skill-based baselines~\citep{li-etal-2026-skillsbench,zhang-etal-2026-coevoskills}. \textbf{Self-Generated} produces one to five emotional support skills in a single pass before interaction, without further refinement. \textbf{CoT-Guided Self-Gen} extends this setting with a structured five-step chain-of-thought prompt. \textbf{SkillCreator} uses Anthropic's Skill Creator framework~\cite{anthropic-2025-skill} to synthesize reusable task instructions from interaction examples. \textbf{HumanCurated} consists of manually designed emotional support skills based on counseling principles and ESC strategy taxonomies.

\paragraph{Metrics.} For response-level evaluation, we report strategy prediction accuracy (ACC), BLEU-1/2/4 (B-1/2/4)~\cite{papineni-etal-2002-bleu}, ROUGE-1/2/L (R-1/2/L)~\cite{lin-2004-rouge}, METEOR (Met)~\cite{banerjee-etal-2005-meteor}, and BERTScore (BS)~\cite{zhang-etal-2019-bertscore}. For dialogue-level evaluation under SAGE, we report the average sentient score (Avg. Score), together with the number of dialogues whose final emotional state exceeds 100 (Success) or falls below 10 (Failure).


\subsection{Experimental Results} 
\subsubsection{Main Results}
Table~\ref{tbl:main_result} shows the results on both the ESConv test set and SAGE benchmark. Detailed results are presented in Table~\ref{tbl:result_escov} and Table~\ref{tbl:result_sage} in Appendix~\ref{apx:performance}, while Appendix~\ref{apx:case-study} presents a case study. 

\begin{table*}[!ht]
\centering
\small
\begin{tabular}{l|lllll|lll}
\toprule
\multirow{2}{*}{\bf Model} & \multicolumn{5}{c|}{\bf ESConv} & \multicolumn{3}{c}{\bf SAGE}\\
& \bf ACC & \bf B-4 & \bf R-L & \bf METEOR & \bf BT & \bf Avg. Score & \bf Success & \bf Failure \\
\midrule
Qwen3.6-Plus & 11.50 & 0.58 & 8.92 & 14.96 & 83.73 & \cellcolor{lightred!85}{66.4} & 13 & \cellcolor{lightred!85}{14}\\
$~~~~$+ESC-Skills & \cellcolor{lightred!85}{23.56} & 0.90 & 10.22 & 20.32 & 84.24 & \cellcolor{darkred!85}{72.1} & \cellcolor{darkred!85}{31} & \cellcolor{darkred!85}{12}\\
\midrule
GPT-5.4-0305-Global & 16.14 & 0.68 & 9.94 & 16.52 & 83.56 & 56.9 & 6 & 21\\
$~~~~$+ESC-Skills & 18.17 & 0.70 & 10.04 & 16.63 & 84.17 & 57.4 & 7 & 19\\
\midrule
Gemini-3.1-Flash & 17.60 & 0.80 & 10.01 & 17.35 & 81.68 & 56.2 & 4 & 21\\
$~~~~$+ESC-Skills & 21.92 & \cellcolor{darkred!85}{1.16} & \cellcolor{darkred!85}{11.96} & 18.18 & \cellcolor{darkred!85}{85.13} & 57.6 & 7 & 19\\
\midrule
Claude-Opus-4.6 & 21.21 & 0.86 & 9.97 & 19.71 & 83.81 & 61.2 & 16 & 18\\
$~~~~$+ESC-Skills & \cellcolor{darkred!85}{23.60} & \cellcolor{lightred!85}{0.93} & 10.26 & \cellcolor{darkred!85}{20.41} & 84.26 & 61.8 & \cellcolor{lightred!85}{19} & 21\\
\midrule
Claude-Sonnet-4.6 & 18.35 & 0.77 & 9.90 & 19.34 & 83.99 & 58.2 & 9 & 23\\
$~~~~$+ESC-Skills & 19.46 & 0.84 & \cellcolor{lightred!85}{10.34} & 19.53 & \cellcolor{lightred!85}{84.34} & 63.6 & 11 & 21\\
\midrule
Claude-Haiku-4.5 & 14.99 & 0.62 & 8.16 & 15.41 & 69.13 & 29.7 & 2 & 51\\
$~~~~$+ESC-Skills & 20.74 & 0.83 & 9.91 & 19.77 & 84.03 & 42.3 & 8 & 43\\
\bottomrule
\end{tabular}
\caption{Performance comparison on the ESC test set and the SAGE benchmark. The baselines are No-Skill baselines. Best results are highlighted in \colorbox{darkred!85}{dark red}, while second-best results are highlighted in \colorbox{lightred!85}{light red}.}
\label{tbl:main_result}
\end{table*}

\begin{table*}[!ht]
\centering
\small
\begin{tabular}{l|lllll|lll}
\toprule
\multirow{2}{*}{\bf Model} & \multicolumn{5}{c|}{\bf ESConv} & \multicolumn{3}{c}{\bf SAGE}\\
& \bf ACC & \bf B-4 & \bf R-L & \bf METEOR & \bf BT & \bf Avg. Score & \bf Success & \bf Failure \\
\midrule
Qwen3.6-Plus & 11.50 & 0.58 & 8.92 & \cellcolor{lightred!85}{14.96} & 83.73 & 66.4 & 13 & 14\\
$~~~~$+Self-Generated & 11.53 & 0.59 & 8.93 & 14.71 & 83.72 & 64.9 & 12 & 16\\
$~~~~$+CoT-Guided Self-Gen & \cellcolor{lightred!85}{12.39} & 0.59 & \cellcolor{lightred!85}{9.04} & 14.86 & \cellcolor{lightred!85}{83.80} & 65.6 & \cellcolor{lightred!85}{16} & \cellcolor{lightred!85}{13}\\
$~~~~$+SkillCreator & 11.89 & 0.57 & 8.80 & \cellcolor{lightred!85}{14.96} & 83.57 & \cellcolor{lightred!85}{67.8} & 14 & 16\\
$~~~~$+HumanCurated & 12.07 & \cellcolor{lightred!85}{0.60} & 9.00 & 14.90 & 83.78 & 62.2 & 15 & 19\\
$~~~~$+ESC-Skills & \cellcolor{darkred!85}{23.56} & \cellcolor{darkred!85}{0.90} & \cellcolor{darkred!85}{10.22} & \cellcolor{darkred!85}{20.32} & \cellcolor{darkred!85}{84.24} & \cellcolor{darkred!85}{72.1} & \cellcolor{darkred!85}{31} & \cellcolor{darkred!85}{12}\\
\bottomrule
\end{tabular}
\caption{Performance comparison of different skill-based baselines using Qwen3.6-Plus on the ESConv test set and the SAGE benchmark.}
\label{tbl:result_baselines}
\end{table*}

\paragraph{Performance on ESConv.} ESC-Skills consistently improves all LLM backbones across response-level evaluation metrics. In particular, substantial gains are observed in strategy prediction accuracy, suggesting that the proposed skill framework helps agents generate responses better aligned with appropriate emotional support strategies. For example, Qwen3.6-Plus achieves a 12.06\% improvement in accuracy after incorporating ESC-Skills. Response quality metrics, including BLEU, ROUGE, METEOR, and BERTScore, also generally improve across models, indicating better semantic relevance and consistency with human supportive responses. The improvements are especially notable for relatively weaker models such as Claude-Haiku-4.5, whose BERTScore increases from 69.13 to 84.03. These results suggest that explicit emotional support skills provide effective behavioral guidance beyond the intrinsic capabilities of the underlying LLMs.

\paragraph{Performance on SAGE.} On the dialogue-level SAGE benchmark, ESC-Skills consistently improves the long-horizon support performance of most LLM backbones. Skill augmentation generally increases the average sentient score while also increasing the number of successful dialogues and reducing severe failure cases. For example, Qwen3.6-Plus improves from 66.4 to 72.1 average sentient score, while the number of successful dialogues increases from 13 to 31. Similar improvements are observed for Gemini-3.1-Flash and Claude-Sonnet-4.6, demonstrating that the proposed skills remain effective in long-horizon multi-turn emotional support interactions.

\subsubsection{Comparison to Baselines}
We use Qwen3.6-Plus as a representative backbone to compare different baselines. As shown in Table~\ref{tbl:result_baselines}, the skill-based baselines provide only marginal improvements or even slightly hurt over the No-Skill setting, while ESC-Skills achieves more gains across both response-level and dialogue-level evaluation. We also find that ESC-Skills outperform human-written ones, consistent with CoEvoSkills~\cite{zhang-etal-2026-coevoskills}.

\paragraph{Comparison on ESConv.} On the ESConv test set, {Self-Generated} and {CoT-Guided Self-Gen} yield only marginal improvements, suggesting that one-pass skill generation is insufficient for learning effective emotional support behaviors. {SkillCreator} and {HumanCurated} also show inconsistent gains across metrics. In contrast, ESC-Skills substantially improves all response-level metrics, including ACC (11.50 $\rightarrow$ 23.56), BLEU-4, ROUGE-L, METEOR, and BERTScore, indicating better modeling of fine-grained intervention behaviors.

\paragraph{Comparison on SAGE.} On the SAGE benchmark, baseline improvements remain limited and unstable. For example, HumanCurated slightly increases successful dialogues but also reduces the average sentient score and increases failure cases, indicating limited robustness across seeker profiles. In contrast, ESC-Skills achieves the best dialogue-level performance, improving the average sentient score from 66.4 to 72.1 and increasing successful dialogues from 13 to 31. This suggests that ESC-Skills provides more robust and adaptive support behaviors in long-horizon multi-turn interactions.

\subsection{Ablation Study}

To analyze the contribution of each component in ESC-Skills, we compare four configurations on Qwen3.6-Plus: (1) the base model without skill augmentation, (2) the initial skill bank $\mathcal{B}^{0}$ constructed from ESConv, (3) $\mathcal{B}^{\triangleright}$, where skills are updated through interaction analysis but without verification, and (4) the final refined bank $\mathcal{B}^{\star}$. Figure~\ref{fig:ablation} summarizes the results, while Table~\ref{tbl:ablation} in Appendix~\ref{apx:performance} reports the full metrics.

\begin{figure*}[t]
    \centering
    \includegraphics[width=0.85\textwidth]{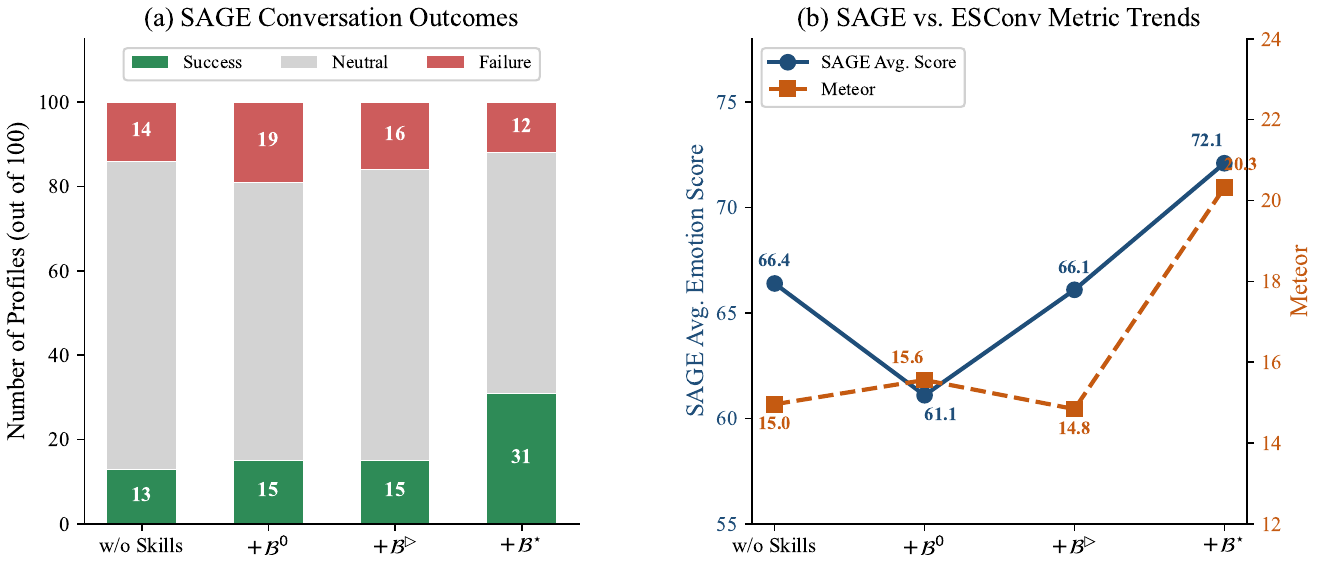}
    \caption{Ablation study on Qwen3.6-Plus. (a)~SAGE conversation outcomes across 100 test profiles\ignore{: $\mathcal{B}^{\star}$ more than doubles the success count while reducing failures}. (b)~Divergent trends between SAGE emotion score and ESConv METEOR\ignore{: $\mathcal{B}^{0}$ improves METEOR but degrades SAGE, while $\mathcal{B}^{\star}$ lifts both}.}
    \label{fig:ablation}
\end{figure*}

\paragraph{Static skills can hurt dynamic performance.} Using the initial skill bank $\mathcal{B}^{0}$ slightly improves several response-level metrics on ESConv, including ROUGE-L and METEOR, indicating better alignment with supportive response patterns. However, dialogue-level performance on SAGE decreases, with lower average emotion scores (66.4 $\rightarrow$ 61.1) and more failure cases (14 $\rightarrow$ 19). We attribute this to a mismatch between static skill instructions and the dynamic nature of long-horizon emotional support interactions. Although the induced skills improve surface-level response quality, some rigid intervention patterns may reduce the agent's flexibility in adapting to evolving emotional cues. This finding suggests that \textit{skill induction from static corpora alone, without interaction-based validation, is insufficient for robust emotional support}.

\paragraph{Evolution without verification is insufficient.}
Updating skills through interaction analysis without verification ($\mathcal{B}^{\triangleright}$) partially recovers the dialogue-level degradation introduced by $\mathcal{B}^{0}$ and slightly reduces failure cases. However, the improvements over the no-skill baseline remain limited on both ESConv and SAGE, indicating that \textit{interaction-driven refinement alone does not reliably produce effective skills}.

\paragraph{Generation–verification refinement loop is the decisive factor.}
The final refined bank $\mathcal{B}^{\star}$ achieves the best overall performance across both benchmarks. On SAGE, it substantially improves average emotion scores and more than doubles the number of successful dialogues compared with the no-skill baseline. It also consistently improves response-level metrics on ESConv, including ACC and METEOR. These results demonstrate that \textit{simulation-based verification plays a critical role in filtering ineffective skills and improving the robustness of emotional support interventions}.


\begin{table}[!ht]
\centering
\footnotesize
\setlength{\tabcolsep}{3pt}
\resizebox{\columnwidth}{!}{%
\begin{tabular}{lcccccccc}
\toprule
\multirow{2}{*}{\textbf{Model}}
& \multicolumn{2}{c}{\textbf{Auto}}
& \multicolumn{3}{c}{\textbf{GPT-Judge}}
& \multicolumn{3}{c}{\textbf{Human}} \\
\cmidrule(lr){2-3}\cmidrule(lr){4-6}\cmidrule(lr){7-9}
& ACC & MET. & Emp. & Help. & Ovrl.
& Emp. & Help. & Ovrl. \\
\midrule
Qwen3.6-Plus
& 11.50 & 14.96 & 4.57 & 3.65 & 4.24
& 4.41 & 3.52 & 4.11 \\
\quad+ESC-Skills
& \textbf{23.56} & \textbf{20.32} & 4.49 & 3.67 & 4.25
& \textbf{4.44} & \textbf{3.69} & \textbf{4.22} \\
\midrule
GPT-5.4-0305
& 16.14 & 16.52 & 4.43 & 3.79 & 4.33
& 4.35 & 3.71 & 4.21 \\
\quad+ESC-Skills
& \textbf{18.17} & \textbf{16.63} & 4.41 & \textbf{3.93} & \textbf{4.42}
& 4.38 & \textbf{3.84} & \textbf{4.31} \\
\midrule
Claude-Haiku-4.5
& 14.99 & 15.41 & 3.68 & 3.27 & 3.55
& 3.55 & 3.14 & 3.47 \\
\quad+ESC-Skills
& \textbf{20.74} & \textbf{19.77} & \textbf{4.05} & \textbf{3.69} & \textbf{4.00}
& \textbf{3.97} & \textbf{3.61} & \textbf{3.91} \\
\bottomrule
\end{tabular}%
}
\caption{Multi-faceted evaluation on three representative model pairs. GPT-Judge scores are generated by GPT-5.4, while human scores are averaged over three annotators on the same 100 ESConv test instances.}
\label{tbl:human-eval}
\end{table}

\subsection{GPT-Judge and Human Evaluation}

Besides automatic metrics, we additionally conduct GPT-Judge and human evaluation on 100 sampled ESConv test instances. We use GPT-5.4 as the judge model and recruit three annotators to evaluate model responses on \textit{Empathy}, \textit{Helpfulness}, and \textit{Overall quality} using a 1--5 Likert scale~\cite{Joshi_Kale_Likert_scale_2015}, with full dialogue context provided.

As shown in Table~\ref{tbl:human-eval}, human evaluation is largely consistent with both automatic metrics and GPT-Judge results across all model pairs. ESC-Skills consistently improves human ratings, with the largest gain observed for Claude-Haiku-4.5 ($\Delta$Overall = +0.44), while GPT-5.4-0305 and Qwen3.6-Plus also achieve smaller but positive improvements. GPT-Judge scores show similar trends, particularly on \textit{Helpfulness} and \textit{Overall quality}. In addition, Fleiss' $\kappa = 0.54$ indicates moderate inter-annotator agreement, while quadratic weighted Cohen's $\kappa_w = 0.65$ between human Overall ratings and GPT-Judge scores suggests substantial alignment, supporting GPT-Judge as a reliable proxy for large-scale evaluation.

\section{Conclusion}
We have proposed \textbf{ESC-Skills}, a skill-centric framework for emotional support conversation that models ESC as an intervention-driven interaction process. By introducing \textbf{Intervention Units (IUs)} to capture localized state--action--outcome dynamics, we construct an executable \textbf{ESC-Skills Bank} from both successful and failed ESC dialogues, and further refine it through multi-profile interaction-based verification under the SAGE framework. Experimental results on both ESConv and SAGE demonstrate that ESC-Skills consistently improves response-level quality and long-horizon emotional support outcomes across multiple LLM backbones. These findings highlight the importance of skill-centric and interaction-driven approaches for building more robust emotional support agents.

\section*{Limitations}
\paragraph{Evaluation scope.}
Following standard practice in ESC research, we evaluate with SAGE, a reproducible simulated help-seeker, rather than live user studies. While SAGE provides controlled, large-scale comparison across conditions, it does not capture the full variability of real human emotional responses. Complementary human evaluation with trained counselors is a natural next step.

\paragraph{Domain and language coverage.}
The current instantiation of ESC-Skills targets English-language emotional support counseling. The evolution framework itself is domain-agnostic, but we have not yet validated it on other supportive dialogue settings (e.g., peer health support, multilingual scenarios). Extending $\mathcal{B}^{\star}$ to additional domains and languages is straightforward in principle and planned as future work.

\paragraph{Skill review.}
Our pipeline automates skill generation and verification through generation–verification refinement loop feedback. In the current version, we do not include a human expert review stage. For deployment in clinical or high-risk settings, integrating licensed counselor oversight into the evolution loop would provide an additional safety layer.

\paragraph{Base model requirements.}
We demonstrate ESC-Skills on strong instruction-following LLMs. Investigating how evolved skills transfer to smaller or open-weight models, and whether skill complexity should adapt to model capacity, remains an open and interesting direction.

\paragraph{Online adaptation.}
The evolved skill bank $\mathcal{B}^{\star}$ is a fixed artifact at deployment time. Enabling continuous, safe online evolution that updates skills from live interaction signals without regression is a promising but non-trivial extension that we leave for future work.

\section*{Ethics Statement}

\paragraph{Data.}
We use the publicly released ESConv corpus~\citep{liu-etal-2021-esconv} under its
original research-use license. The corpus contains anonymized peer-support
dialogues; no additional personally identifiable information is collected
or released by this work.

\paragraph{Intended use and risks.}
ESC-Skills is designed as a \emph{research artifact} to study skill-based
emotional support, \textbf{not as a substitute for licensed mental-health
professionals}. The system must not be deployed in crisis-intervention or
clinical-decision pipelines without expert oversight and rigorous safety
auditing. Generated responses may occasionally fail to recognize crisis
signals; downstream applications must integrate dedicated safety classifiers
and human escalation paths.

\paragraph{Human annotation.}
Three annotators (proficient in English, holding at least a bachelor's degree)
were recruited for the human evaluation. Annotators were informed in advance
of the emotionally sensitive content, given the option to opt out at any time,
and provided with mental-health resource references. Compensation was set
above the local minimum wage. Detailed guidelines are provided in
Appendix.

\paragraph{LLM usage.}
We rely on third-party LLM APIs for both the agent and the judge. All API calls
comply with the providers' terms of service. No model weights are released;
only the skill bank ($\mathcal{B}^{\star}$) and evaluation code will be
made public.

\paragraph{Reproducibility.}
All prompts (Appendix~\ref{apx:prompts}), evolution hyperparameters, sampled turn indices, and aggregated judge scores will be released to support reproduction without re-disclosing raw seeker utterances beyond what is already public in ESConv.



\bibliography{custom}


\appendix

\section{Intervention Unit Annotation Details}\label{apx:iu_annotation}

To construct Intervention Units (IUs), we design a multi-dimensional annotation framework covering dialogue-level scenarios, seeker emotional states, supporter intervention actions, and post-intervention response changes. The label sets are iteratively developed through manual inspection of ESConv and FailedESConv conversations together with preliminary LLM-based open coding. Specifically, we first sample representative conversations from both successful and failed ESC interactions, identify recurring emotional situations and intervention behaviors, and then consolidate semantically overlapping categories into a unified taxonomy. The resulting labels are designed to balance coverage, interpretability, and annotation consistency while remaining sufficiently fine-grained for modeling localized intervention dynamics.

Table~\ref{tbl:scenario} lists the 18 dialogue-level scenario labels used to characterize the seeker's real-world situation. Table~\ref{tbl:state} presents the 15 seeker emotional states used for utterance-level state annotation. Table~\ref{tbl:action} shows the 17 supporter intervention actions used to describe intervention behaviors, which extend the original ESConv strategy taxonomy with more fine-grained and intervention-oriented categories. Table~\ref{tbl:change} lists the 14 seeker response-change labels used to characterize post-intervention emotional transitions.

During annotation, Claude-Opus is prompted to jointly analyze the dialogue context, seeker utterances, supporter responses, and subsequent emotional reactions in order to produce structured annotations under the predefined label sets. Dialogue-level scenario labels are assigned at the conversation level, while seeker states, support actions, and response changes are annotated at the utterance level. The resulting annotations are then used to construct localized Intervention Units (IUs) for subsequent skill induction and refinement.

\begin{table*}[!t]
\centering
\scriptsize
\resizebox{\linewidth}{!}{
\begin{tabular}{p{0.28\linewidth}|p{0.64\linewidth}}
\toprule
\bf Scenario Label & \bf Description \\
\midrule
Loss of perceived control & The seeker feels unable to influence life events, emotions, or ongoing situations. \\
Anxiety and stress & The seeker experiences persistent worry, tension, or stress-related burden. \\
Loneliness & The seeker feels emotionally isolated, disconnected, or lacking companionship. \\
Doubts about self-worth & The seeker questions their own value, adequacy, or deservingness. \\
Loss and grief & The seeker is coping with bereavement, separation, or another meaningful loss. \\
Trust rupture & The seeker experiences betrayal or broken trust in an important relationship. \\
Career uncertainty / intimate relationship conflict & The seeker feels unsure about career direction or distressed by romantic conflict. \\
Excessive sense of responsibility & The seeker feels overly responsible for others or for keeping things stable. \\
Feelings of neglect & The seeker feels overlooked, uncared for, or insufficiently attended to. \\
Family conflict & The seeker is experiencing persistent tension or disagreement within the family. \\
Social withdrawal & The seeker withdraws from social interaction or avoids interpersonal contact. \\
Depressed mood & The seeker presents sadness, heaviness, or a sustained low mood state. \\
Interpersonal conflict & The seeker is involved in conflict or relational difficulty with others. \\
Self-negation & The seeker dismisses or suppresses their own needs, feelings, or identity. \\
Perfectionism-related distress & The seeker feels distress driven by unrealistic standards or fear of mistakes. \\
Impaired personal boundaries & The seeker has difficulty establishing or maintaining healthy boundaries. \\
Identity confusion & The seeker feels uncertain or conflicted about their sense of self. \\
\bottomrule
\end{tabular}
}
\caption{List of the 18 scenario labels used in the dialogue-level annotation.}
\label{tbl:scenario}
\end{table*}

\begin{table*}[!t]
\centering
\scriptsize
\resizebox{\linewidth}{!}{
\begin{tabular}{p{0.28\linewidth}|p{0.64\linewidth}}
\toprule
\bf Emotional State & \bf Description \\
\midrule
Willingness to explore & The seeker shows openness to discussing their experience in greater depth. \\
Self-awareness & The seeker demonstrates insight into their emotions, thoughts, or behavioral patterns. \\
Depressed mood & The seeker expresses sadness, heaviness, or a sustained low emotional state. \\
Intellectualization & The seeker focuses on analysis or reasoning while distancing from emotional experience. \\
Helplessness & The seeker feels powerless, stuck, or unable to change the situation. \\
Advice seeking & The seeker explicitly asks for guidance, direction, or practical suggestions. \\
Tentative disclosure & The seeker shares cautiously, indirectly, or with hesitation. \\
Heightened emotional arousal & The seeker is emotionally activated, intense, or overwhelmed in the moment. \\
Rumination & The seeker repeatedly circles around the same thoughts, concerns, or dilemmas. \\
Disorganized expression & The seeker’s expression is fragmented, unclear, or difficult to follow. \\
Avoidance & The seeker evades difficult topics, emotions, or direct engagement. \\
Indecisiveness & The seeker struggles to make choices or commit to a direction. \\
Anger expression & The seeker expresses anger, frustration, or resentment intensely. \\
Self-blame & The seeker attributes excessive fault or responsibility to themselves. \\
High defensiveness & The seeker shows resistance, guardedness, or strong self-protective responses. \\
\bottomrule
\end{tabular}
}
\caption{List of the 15 seeker emotional states used in the utterance-level annotation.}
\label{tbl:state}
\end{table*}

\begin{table*}[!t]
\centering
\scriptsize
\resizebox{\linewidth}{!}{
\begin{tabular}{p{0.28\linewidth}|p{0.64\linewidth}}
\toprule
\bf Supporter Action & \bf Description \\
\midrule
Action-oriented suggestions & The supporter offers concrete steps or practical recommendations for coping or problem-solving. \\
Strengths/resource affirmation & The supporter highlights the seeker’s strengths, coping capacities, or available resources. \\
Open-ended questioning & The supporter invites elaboration through broad questions that encourage further sharing. \\
Empathic reflection & The supporter reflects the seeker’s feelings or experience in an understanding and validating way. \\
Supporter self-disclosure & The supporter shares personal experience or perspective to build connection or normalize experience. \\
Information provision & The supporter provides relevant knowledge, explanations, or psychoeducational content. \\
Normalization & The supporter conveys that the seeker’s reactions are understandable or common under the circumstances. \\
Closed-ended questioning & The supporter asks focused questions that can be answered briefly or specifically. \\
Cognitive reframing & The supporter offers an alternative interpretation to help the seeker view the situation differently. \\
Exploratory deepening & The supporter encourages deeper examination of underlying feelings, meanings, or patterns. \\
Paraphrasing and clarification & The supporter restates the seeker’s message to confirm understanding or reduce ambiguity. \\
Boundary setting/reminder & The supporter reinforces interpersonal limits, roles, or appropriate relational boundaries. \\
Emotion labeling & The supporter explicitly names or identifies the seeker’s emotional state. \\
Guided questioning & The supporter uses directional questions to help the seeker reflect in a structured way. \\
Summarizing and focusing & The supporter synthesizes key points and helps concentrate the conversation on central issues. \\
Gentle challenge & The supporter respectfully questions inconsistencies, assumptions, or unhelpful patterns. \\
Intentional silence & The supporter allows space and pause for emotional processing or continued disclosure. \\
\bottomrule
\end{tabular}
}
\caption{List of the 17 supporter intervention actions used in the utterance-level annotation.}
\label{tbl:action}
\end{table*}

\begin{table*}[!t]
\centering
\scriptsize
\setlength{\tabcolsep}{4pt}
\begin{tabular}{p{0.28\linewidth}|p{0.64\linewidth}}
\toprule
\bf Response Change & \bf Description \\
\midrule
More specific expression & The seeker provides more concrete, detailed, or precise descriptions than before. \\
Continued disclosure & The seeker continues sharing thoughts, feelings, or experiences without shutting down. \\
No observable change & The seeker’s response shows no clear shift in emotion, engagement, or direction. \\
Emotional relief & The seeker appears calmer, less distressed, or emotionally eased after the response. \\
Expression of willingness to take action & The seeker indicates readiness to try a coping step or make a change. \\
Willingness to consider a new perspective & The seeker shows openness to viewing the situation from a different angle. \\
Indeterminable & The response does not provide enough information to infer a clear change. \\
Topic shift & The seeker moves the conversation away from the current issue to a different topic. \\
Increased self-awareness & The seeker begins to notice or articulate internal patterns, emotions, or motives. \\
Increased confusion & The seeker appears more uncertain, disorganized, or unclear than before. \\
Increased withdrawal & The seeker becomes more closed, distant, or less willing to engage. \\
Increased emotional agitation & The seeker becomes more emotionally activated, upset, or escalated. \\
Reduced repetitive responding & The seeker stops repeating the same statements or thought patterns. \\
Perceived offense & The seeker appears to feel hurt, misunderstood, or offended by the response. \\
\bottomrule
\end{tabular}
\caption{List of the 14 seeker response-change labels used in the utterance-level annotation.}
\label{tbl:change}
\end{table*}

\section{More Details of the Skill Prototypes}\label{apx:prototype}

\ignore{
\begin{figure}[!t]
\centering
\includegraphics[width=\columnwidth, trim={0cm 0cm 0cm 0cm}]{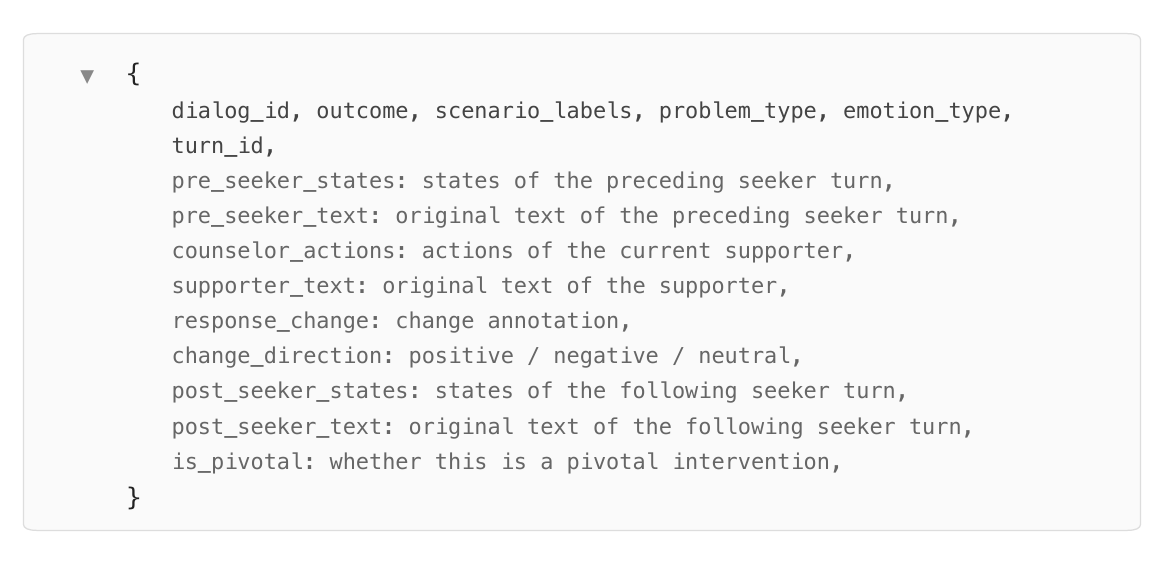}
\caption{Detail of an intervention unit.}
\label{fig:iu}
\end{figure}
}

\begin{table}[t]
\centering
\small
\begin{tabular}{p{0.32\linewidth} p{0.60\linewidth}}
\toprule
\textbf{Field} & \textbf{Description} \\
\midrule
{dialog\_id} & Dialogue identifier \\
{outcome} & Overall conversation outcome \\
{scenario\_labels} & Dialogue-level scenario labels \\
{problem\_type} & Type of emotional problem \\
{emotion\_type} & Primary emotional category \\
{turn\_id} & Dialogue turn index \\
{pre\_seeker\_states} & Emotional states before intervention \\
{pre\_seeker\_text} & Seeker utterance before intervention \\
{counselor\_actions} & Support intervention actions \\
{supporter\_text} & Supporter response text \\
{response\_change} & Post-response emotional change \\
{change\_direction} & Positive / negative / neutral change \\
{post\_seeker\_states} & Emotional states after intervention \\
{post\_seeker\_text} & Seeker utterance after intervention \\
{is\_pivotal} & Whether the intervention is pivotal \\
\bottomrule
\end{tabular}
\caption{Structure of an Intervention Unit (IU). Each IU records localized state--action--outcome interaction dynamics surrounding a supporter intervention.}
\label{tbl:iu_structure}
\end{table}

For each group (prototype), we compute its \textit{effectiveness rate}, defined as the proportion of positive IUs within the group. Table~\ref{tbl:prototype1} presents eight representative skill prototypes with a 100.0\% effectiveness rate, indicating that all associated IUs lead to positive post-response emotional changes. These examples suggest that certain support actions consistently produce constructive intervention outcomes when applied under specific seeker states.

Table~\ref{tbl:prototype2} shows examples of filtered skill prototypes whose effectiveness rates fall below the predefined threshold. Interestingly, these cases reveal that even when the same support action is applied to the same seeker state, the resulting emotional outcomes may vary substantially across conversations. In particular, some interventions may simultaneously exhibit constructive effects in certain contexts while causing negative impacts such as increased emotional agitation, withdrawal, confusion, or feelings of invalidation in others. These observations further highlight the importance of modeling intervention effectiveness and risk sensitivity in emotional support conversations.

\begin{table*}[t]
\centering
\scriptsize
\setlength{\tabcolsep}{3pt}
\begin{tabular}{p{0.21\linewidth} p{0.18\linewidth} c p{0.40\linewidth}}
\toprule
\bf Seeker State & \bf Support Action & \bf \#IUs & \bf Description \\
\midrule
Self-awareness & Open-ended questioning & 238 & Helps reflective seekers elaborate. \\
Self-awareness & Exploratory deepening & 185 & Supports deeper exploration of inner experience. \\
Intellectualization & Open-ended questioning & 140 & Moves the seeker beyond abstract analysis. \\
Indecisiveness & Information provision & 88 & Provides concrete information to reduce uncertainty. \\
Indecisiveness & Normalization & 74 & Frames hesitation as understandable. \\
Self-awareness & Summarizing and focusing & 74 & Synthesizes reflections around the core issue. \\
Indecisiveness & Gentle challenge & 23 & Surfaces avoidance or conflicting assumptions. \\
Self-blame & Boundary setting/reminder & 9 & Reduces excessive self-blame via boundary clarification. \\
\bottomrule
\end{tabular}
\caption{Eight skill prototypes achieving an effectiveness rate of 1.0.}
\label{tbl:prototype1}
\end{table*}

\begin{table*}[t]
\centering
\scriptsize
\setlength{\tabcolsep}{3pt}
\begin{tabular}{p{0.25\linewidth} p{0.25\linewidth} c p{0.23\linewidth}}
\toprule
\bf Seeker State & \bf Support Action & \bf Eff. Rate & \bf Negative Impact \\
\midrule
High defensiveness & Boundary setting/reminder & 42.9\% & Increased emotional agitation \\
High defensiveness & Cognitive reframing & 47.6\% & Increased emotional agitation \\
Disorganized expression & Boundary setting/reminder & 50.0\% & Increased withdrawal \\
Disorganized expression & Gentle challenge & 50.0\% & Increased confusion \\
High defensiveness & Gentle challenge & 57.1\% & Perceived offense \\
\bottomrule
\end{tabular}
\caption{Examples of prototypes filtered out due to effectiveness rates below the threshold.}
\label{tbl:prototype2}
\end{table*}

\section{Example of Emotional Support Skill}
\label{apx:skill}

To make the design of $\mathcal{B}^{\star}$ concrete, we reproduce one of the two skills that fire in the \textsc{+ESC-Skills} arm on the case study in Appendix~\ref{apx:case-study}: \texttt{esc-strategy-switching}. The full \texttt{SKILL.md} is 461 lines and 8 sections; below we keep the YAML frontmatter, every \texttt{H1/H2/H3} heading, and one representative bullet per section, replacing the remainder with ``\texttt{[...]}'' so the artefact fits on a single page. The companion skill, \texttt{esc-action-planning}, follows the same structure.

\begin{figure*}[h]
\centering
\begin{lstlisting}[
  basicstyle=\scriptsize\ttfamily,
  breaklines=true,
  columns=fullflexible,
  keepspaces=true,
  frame=single,
  framesep=4pt,
  xleftmargin=6pt,
  xrightmargin=6pt,
  aboveskip=4pt,
  belowskip=4pt,
]
---
name: esc-strategy-switching
description: Decide when and how to switch intervention strategies mid-conversation. Use when the current approach is not working, the seeker's state has changed, or a transition between conversation phases is needed. Includes urgent override triggers for repeated unmet advice-seeking bids, [...].
metadata:
  domain: emotional-support-counseling
  category: meta
  version: "3.2"
---

# Dynamic Strategy Switching

## Overview
A meta-skill that continuously assesses whether the current intervention still matches the client's state, and switches promptly when it does not. It addresses three problem types:
(1) the current strategy is ineffective,
(2) the client's state has changed, 
(3) the conversation phase needs to shift. This is a higher-order decision mechanism, not a standalone response technique. [...]

### Critical Failure Modes This Skill Must Prevent
#### Failure Mode #1: Endless Reflection When Advice Is Needed
Staying in reflective/empathic mode for 5--7+ turns after the seeker has explicitly and repeatedly signaled a need for practical guidance. This leads to declining seeker emotion, frustration, and loss of alliance. 
[... Failure Modes #2--#8 omitted: Poetic Over-Elaboration,
   Ignoring Communication Style, Long First Replies, Metaphor-Mirroring
   Trap, Validating Without Advancing, Ignoring Explicit "How" Questions,
   Refusing to Shift Register for Entrenched Seekers ...]

## Mandatory Decision Rules (Execute Before Drafting Any Response)

### Rule 1: The "How" Question Override (HIGHEST PRIORITY)
- Trigger: seeker asks "how do I," "what should I do," "is there a way," "what would help," "where do I start," "do you have suggestions" ...
- Action: lead with concrete, specific, actionable guidance (numbered steps, bullets, or specific examples). Emotional acknowledgment may follow, never precede.
- Forbidden: starting with "That question is so [adjective]...". ANSWER the question.
[... Rules 0, 2--7 omitted: First Response Protocol, Metaphor Escape Rule,
   Stagnation Counter, Advice-Seeking Escalation Ladder, Entrenched
   Intellectual Override, Response Length & Energy Matching, Poetry
   Budget ...]

## Core Principles

### 0. Respond to the Need, Not the Aesthetic Before drafting any response, ask: 
(1) What is the seeker actually asking for? 
(2) Have I already provided that in my last 1--2 responses?
(3) Am I about to extend their metaphor instead of addressing the problem? [...]
[... Principles 1--5 omitted: state-action matching with effectiveness
   table, outcome-signal monitoring, repetitive-loop detection,
   register-matching, the Advancement Principle ...]

## Specific Scenario Guidance
[... 4 scenarios omitted: Parent-Child Career Conflict, Creative Block
   with Deadline, Long-Distance Relationship Anxiety, Gentle/Tentative
   Seeker with Persistent Anxiety ...]

## Response Construction Protocol (step-by-step for every response)
Step 1: Classify the seeker's current bid.
Step 2: Check the stagnation counter.
Step 3: Check the poetry budget.
Step 4: Check the advancement criterion.
Step 5: Check length.
Step 6: Draft and revise.
[...]

## Anti-Pattern Library
[... extensive examples extracted from failed evolution rounds,
   including poetic over-elaboration, metaphor mirroring, validation
   loops, and intellectual-mismatch transcripts ...]
\end{lstlisting}
\caption{An abridged \texttt{SKILL.md} for \texttt{esc-strategy-switching},
one of the 34 skills in $\mathcal{B}^{\star}$. The YAML frontmatter and
the H1/H2/H3 heading hierarchy are preserved verbatim; bulk prose is
replaced with ``\texttt{[...]}'' so the example fits one page. This is
the skill whose ``Rule~1: The `How' Question Override'' fires on the
seeker's confirmation cue in Appendix~\ref{apx:case-study}, after which the
companion skill \texttt{esc-action-planning} produces the concrete plan.}
\label{fig:skill-example}
\end{figure*}

\section{Composition of the Final Skill Bank $\mathcal{B}^{\star}$}
\label{apx:bank-composition}

Table~\ref{tbl:bank-composition} lists all 34 skills constituting $\mathcal{B}^{\star}$, grouped by their \texttt{metadata.category}. \textbf{Origin} marks each skill as inherited from $\mathcal{B}^{0}$, rewritten by the evolution (\textit{updated}), or newly introduced (\textit{added}).

\begin{table*}[t]
\centering
\scriptsize
\renewcommand{\arraystretch}{1.0}
\setlength{\tabcolsep}{3pt}
\begin{tabularx}{\textwidth}{@{} l l X @{}}
\toprule
\textbf{Skill} & \textbf{Origin} & \textbf{Description} \\
\midrule
\multicolumn{3}{l}{\textbf{Meta (orchestration \& safety)} (4 skills)} \\
\midrule
\texttt{esc-failure-recovery} & $\mathcal{B}^{0}$ & Recover from failed interventions when the seeker becomes more distressed, defensive, or disengaged after a response. Use immedia… \\
\texttt{esc-risk-awareness} & $\mathcal{B}^{0}$ & Critical safety guidelines for emotional support. ALWAYS check this skill before using confrontational techniques. Contains contr… \\
\texttt{esc-state-assessment} & $\mathcal{B}^{0}$ (updated) & Real-time assessment of seeker's emotional and cognitive state. Use continuously throughout the conversation to identify the seek… \\
\texttt{esc-strategy-switching} & $\mathcal{B}^{0}$ (updated) & Decide when and how to switch intervention strategies mid-conversation. Use when the current approach is not working, the seeker'… \\
\midrule
\multicolumn{3}{l}{\textbf{Conversation Phase} (4 skills)} \\
\midrule
\texttt{esc-closing-consolidation} & $\mathcal{B}^{0}$ & Consolidate gains and plan next steps in the closing phase. Use when wrapping up the conversation to summarize insights, affirm r… \\
\texttt{esc-intervention-deepening} & $\mathcal{B}^{0}$ & Deepen intervention and facilitate change in the middle-late phase. Use when the problem is understood and the seeker is ready fo… \\
\texttt{esc-opening-rapport} & $\mathcal{B}^{0}$ (updated) & Establish rapport and safety in the opening phase of emotional support conversations. Use at the start of any new conversation to… \\
\texttt{esc-problem-exploration} & $\mathcal{B}^{0}$ & Explore and assess the seeker's core concerns in the early-middle phase. Use after initial rapport to help the seeker articulate … \\
\midrule
\multicolumn{3}{l}{\textbf{Technique} (12 skills)} \\
\midrule
\texttt{esc-action-planning} & $\mathcal{B}^{0}$ (updated) & Guide concrete action planning and provide relevant information. Use in later conversation phases when the seeker is ready for pr… \\
\texttt{esc-advice-readiness-detection} & added & Detect when a seeker is explicitly or implicitly requesting practical advice—through direct questions, growing frustration with r… \\
\texttt{esc-authentic-attunement} & added & Adapt empathic responses to match the seeker's emotional register, communication style, and depth needs—avoiding over-polished or… \\
\texttt{esc-cognitive-reframing} & $\mathcal{B}^{0}$ & Apply cognitive reframing and gentle challenge techniques safely. Use when the seeker is ready for perspective shifts. WARNING: h… \\
\texttt{esc-dialectical-analysis} & added & Engage seekers in balanced, multi-perspective examination of their situation by exploring tensions, contradictions, competing val… \\
\texttt{esc-empathic-reflection} & $\mathcal{B}^{0}$ (updated) & Master empathic reflection including emotional naming and paraphrasing. Use as the foundational technique throughout conversation… \\
\texttt{esc-exploration-questioning} & $\mathcal{B}^{0}$ & Use different questioning techniques strategically. Covers open, closed, guided, and deepening questions. Use to help the seeker … \\
\texttt{esc-motive-perspective-analysis} & added & Help seekers collaboratively analyze why another person may be acting as they are by generating grounded, uncertainty-aware hypot… \\
\texttt{esc-normalization-validation} & $\mathcal{B}^{0}$ (updated) & Apply normalization and resource affirmation. Use to reduce shame, validate the seeker's reactions as understandable, and highlig… \\
\texttt{esc-other-perspective-analysis} & added & Help seekers collaboratively analyze and understand other people's motivations, behavioral patterns, and psychological drivers us… \\
\texttt{esc-specific-effort-recognition} & added & Recognize covert bids for acknowledgment and respond with sincere, concrete praise of the seeker's specific actions, effort, rest… \\
\texttt{esc-unfair-blame-validation} & added & Support seekers who feel wrongly blamed or seek exoneration by validating unfairness and the need to be understood while avoiding… \\
\midrule
\multicolumn{3}{l}{\textbf{Scenario \& Seeker State} (14 skills)} \\
\midrule
\texttt{esc-ambivalence-guidance} & $\mathcal{B}^{0}$ (updated) & Guide ambivalent seekers toward clarity and action. Use when the seeker is indecisive, torn between options, or explicitly asking… \\
\texttt{esc-anxiety-overwhelm} & $\mathcal{B}^{0}$ & Help seekers overwhelmed by anxiety and loss of control. Use when the seeker reports persistent worry, feeling things are spirali… \\
\texttt{esc-boundary-overload} & $\mathcal{B}^{0}$ & Help seekers who over-extend themselves or withdraw socially. Use when the seeker takes on too much responsibility, cannot set bo… \\
\texttt{esc-career-uncertainty} & $\mathcal{B}^{0}$ & Support seekers facing career uncertainty, job dissatisfaction, or perfectionism pressure. Use when the seeker is struggling with… \\
\texttt{esc-confusion-clarification} & $\mathcal{B}^{0}$ & Help seekers who are confused or caught in repetitive thinking loops. Use when the seeker's expression is disorganized, they keep… \\
\texttt{esc-emotional-crisis} & $\mathcal{B}^{0}$ & Handle emotional crises in support conversations. Use when the seeker shows intense emotional activation, anger outbursts, or is … \\
\texttt{esc-grief-and-loss} & $\mathcal{B}^{0}$ & Support seekers experiencing grief, loneliness, or feeling overlooked. Use when the seeker has lost someone or something importan… \\
\texttt{esc-insight-deepening} & $\mathcal{B}^{0}$ (updated) & Deepen self-awareness and exploration when the seeker is open. Use when the seeker shows readiness to explore their feelings, beg… \\
\texttt{esc-intellectualization-grounding} & $\mathcal{B}^{0}$ & Ground intellectualizing seekers in emotional experience. Use when the seeker talks about feelings abstractly, uses distancing la… \\
\texttt{esc-low-mood-support} & $\mathcal{B}^{0}$ & Support seekers experiencing low mood, hopelessness, or helplessness. Use when the seeker expresses sadness, feeling stuck, or a … \\
\texttt{esc-relationship-conflict} & $\mathcal{B}^{0}$ (updated) & Navigate relationship conflicts including intimate partner disputes, family tensions, interpersonal conflicts, and trust betrayal… \\
\texttt{esc-resistance-handling} & $\mathcal{B}^{0}$ & Handle defensive or avoidant seekers in emotional support conversations. Use when the seeker shows resistance, deflects questions… \\
\texttt{esc-self-blame-response} & $\mathcal{B}^{0}$ & Respond to seekers who are self-blaming or experiencing guilt. Use when the seeker attributes problems entirely to themselves, sh… \\
\texttt{esc-self-worth-crisis} & $\mathcal{B}^{0}$ & Address self-worth crises including self-doubt, self-negation, and identity confusion. Use when the seeker questions their fundam… \\
\bottomrule
\end{tabularx}
\caption{Composition of the final skill bank $\mathcal{B}^{\star}$ (34 skills), grouped by functional family. \textbf{Origin}: \textit{inherited}~$\mathcal{B}^{0}$, \textit{updated}~(rewritten by evolution), \textit{added}~(newly introduced).}
\label{tbl:bank-composition}
\end{table*}

\section{Case Study: A Strategy-Switching Failure Mode}
\label{apx:case-study}

To make the aggregate gains in Table ~\ref{tbl:main_result} concrete, we walk through a single ESConv supporter turn on which all six arms produce qualitatively different replies. The full six-arm comparison is shown in Table~\ref{tbl:case-study-full}.

\paragraph{Dialogue context.}
The seeker is a long-term unemployed adult returning to work after a two-year break. Two preceding supporter turns have already provided suggestions (\textit{doing little things every day}'',\textit{really adds up}''). The seeker's last utterance, \textit{OK, so with, well, kinf0f like baby steps\dots right}'', is an explicit confirmation request: the seeker is no longer disclosing distress, but \emph{seeking the next concrete action}. The gold supporter response is consequently labelled \textsc{Providing Suggestions}.

\paragraph{Observation 1 — Five baselines collapse onto the same empathic prior.}
\textsc{Qwen3.6-Plus}, \textsc{+Self-Gen}, \textsc{+CoT-Self-Gen}, \textsc{+SkillCreator}, and \textsc{+HumanCurated} all select \textsc{Reflection of Feelings} and produce near-paraphrastic openings (\textit{It makes complete sense to feel nervous \dots}''). Strikingly, the three skill-construction pipelines (model self-generation, CoT-guided self-generation, and Anthropic's \texttt{skill-creator}) converge on essentially identical text, indicating that none of these methods has overridden the base model's empathy-by-default tendency on this turn.

\paragraph{Observation 2 — $\mathcal{B}^{\star}$ is the only arm that switches strategy.} \textsc{+ESC-Skills} selects \textsc{Providing Suggestions}, matching the gold annotation, and grounds the reply in specific actions (\textit{update your resume or browse one job board for 10 minutes; reach out to one former colleague}''). Inspecting the activated skills reveals a two-step composition: \texttt{esc-strategy-switching} fires on the seeker's confirmation cue, after which \texttt{esc-action-planning} produces the concrete plan. Both skills were synthesised during the SAGE-in-the-loop evolution of $\mathcal{B}^{\star}$ precisely in response to repeated over-empathy failures observed on $\mathcal{B}_0$.

\paragraph{Observation 3 — Qualitative collapse mirrors the quantitative ranking.}
The near-identity of the five non-Ours replies is consistent with their main-table SAGE deltas over Base ($\le!1.5$ points), whereas \textsc{+ESC-Skills} contributes $+8.7$. The case localises where the gap is realised in practice: not in how empathy is \emph{worded}, but in whether the supporter \emph{stops empathising and starts advising} when the seeker signals readiness.

\paragraph{Takeaway.}
Producing fluent ESC skills is necessary but not sufficient. What separates $\mathcal{B}^{\star}$ from the self-generation, CoT-guided, skill-creator, and human-curated banks is coverage of the \emph{strategic transitions} that long-horizon ESC dialogue demands. This case exhibits the simplest such transition --- empathy to action --- and only the SAGE-evolved bank handles it correctly.

\begin{table*}[t]
\centering
\scriptsize
\setlength{\tabcolsep}{3pt}
\renewcommand{\arraystretch}{1.10}
\begin{tabularx}{\textwidth}{@{} l|X|X|X|X|X|X @{}}
\toprule
 & \texttt{Qwen3.6-Plus} & \texttt{+Self-Gen} & \texttt{+CoT-Self-Gen} & \texttt{+SkillCreator} & \texttt{+HumanCurated} & \texttt{+ESC-Skills ($\mathcal{B}^\star$)} \\
\midrule
\textbf{Meta} & \multicolumn{6}{@{}p{\dimexpr\textwidth-3.5em\relax}@{}}{\textbf{Emotion}: \textit{anxiety} \quad \textbf{Problem}: \textit{Procrastination} \quad \textbf{Experience}: \textit{Current Experience} \newline \textbf{Situation}: Applying for work, I have been retired for 2 years and would like to return to the work force.} \\
\midrule
\textbf{Context} & \multicolumn{6}{@{}p{\dimexpr\textwidth-3.5em\relax}@{}}{\textbf{Supporter}: [Providing Suggestions] Doing little things every day \newline \textbf{Seeker}: So, keep myself busy and try to ignore my nervousness? \newline \textbf{Supporter}: [Providing Suggestions] Really adds up and helps to get the big picture moving. \newline \textbf{Seeker}: OK, so with, well, kinf0f like baby steps....right} \\
\midrule
\textbf{Gold} & \multicolumn{6}{@{}p{\dimexpr\textwidth-3.5em\relax}@{}}{[\textsc{Providing Suggestions}] It is about making small adjustments to feed a good habit.} \\
\midrule
\textbf{Reply} & [\textsc{Reflection of feelings} $\times$]\newline It makes complete sense that you'd feel nervous about rejection — especially after being out of the workforce for a while. That worry isn’t a sign that you’re … & [\textsc{Reflection of feelings} $\times$]\newline It makes complete sense to feel nervous—especially when stepping back into something important after time away. That worry about rejection isn’t a sign you’re … & [\textsc{Reflection of feelings} $\times$]\newline It makes complete sense to feel nervous—especially when stepping back into something after a break, and especially when rejection feels so personal. That worry… & [\textsc{Reflection of feelings} $\times$]\newline It makes complete sense that you'd feel nervous about rejection — especially after being out of the workforce for a while. That worry isn’t a sign that you’re … & [\textsc{Reflection of feelings} $\times$]\newline It makes complete sense to feel nervous—especially when stepping back into something after time away. That worry about rejection isn’t a sign you’re not ready;… & [\textsc{Providing Suggestions} $\checkmark$]\newline Yes — baby steps are exactly right. For example: today, you might just update your resume or browse one job board for 10 minutes. Tomorrow, you could reach out… \\
\bottomrule
\end{tabularx}
\caption{Full six-arm comparison on ESConv sample.
The Meta row reproduces the conversation-level annotation from ESConv
(emotion, problem, experience type, and seeker situation).
$\checkmark$/$\times$ indicates whether the predicted strategy matches the gold strategy \textsc{Providing Suggestions}.}
\label{tbl:case-study-full}
\end{table*}

\section{Detailed Performance}~\label{apx:performance}
Table~\ref{tbl:result_escov} shows the detailed performance on the ESConv test set. Table~\ref{tbl:result_sage} shows the detailed performance on the SAGE benchmark with multiple dialogue-level metrics, including average seeker emotion score (Avg.Score), median score (Media), minimum/maximum score (Min/Max), numbers of successful and failed dialogues (Success/Failures), and the distribution of final emotional grades (S/A/B/C/F). Higher Avg.Score, Media, Max, and Success indicate better emotional support quality, while lower Failures indicates fewer harmful interactions. The grade distribution reflects the overall emotional outcomes of dialogues, where S and A denote highly positive outcomes and F denotes severe emotional deterioration.

\begin{table*}[!ht]
\centering
\small
\begin{tabular}{l|l|lll|lll|ll}
\toprule
\bf Model & \bf ACC & \bf B-1 & \bf B-2 & \bf B-4 & \bf R-1 & \bf R-2 & \bf R-L & \bf METEOR & \bf BERTScore\\
\midrule
Qwen3.6-Plus & 11.50 & 8.39 & 2.00 & 0.58 & 12.72 & 1.07 & 8.92 & 14.96 & 83.73 \\
$~~$+ESC-Skills & 23.56 & 9.08 & 3.07 & 0.90 & 14.38 & 2.36 & 10.22 & 20.32 & 84.24\\
\midrule
GPT-5.4-0305-global & 16.14 & 8.85 & 2.41 & 0.68 & 14.06 & 1.38 & 9.94 & 16.52 & 83.56\\
$~~$+ESC-Skills & 18.17 & 9.16 & 2.48 & 0.70 & 14.17 & 1.50 & 10.04 & 16.63 & 84.17\\
\midrule
Gemini-3.1-flash & 17.60 & 9.64 & 2.86 & 0.80 & 14.06 & 1.80 & 10.01 & 17.35 & 81.68\\
$~~$+ESC-Skills & 21.92 & 11.90 & 3.61 & 1.16 & 16.30 & 2.34 & 11.96 & 18.18 & 85.13\\
\midrule
Claude-opus-4.6 & 21.21 & 9.08 & 2.93 & 0.86 & 14.14 & 2.18 & 9.97 & 19.71 & 83.81\\
$~~$+ESC-Skills & 23.60 & 9.18 & 3.11 & 0.93 & 14.46 & 2.40 & 10.26 & 20.41 & 84.26\\
\midrule
Claude-sonnet-4.6 & 18.35 & 9.21 & 2.77 & 0.77 & 14.17 & 1.92 & 9.90 & 19.34 & 83.99\\
$~~$+ESC-Skills & 19.46 & 9.51 & 3.00 & 0.84 & 14.57 & 2.15 & 10.34 & 19.53 & 84.34\\
\midrule
Claude-haiku-4.5 & 14.99 & 7.68 & 2.24 & 0.62 & 11.56 & 1.39 & 8.16 & 15.41 & 69.13\\
$~~$+ESC-Skills & 20.74 & 8.77 & 2.88 & 0.83 & 13.90 & 2.13 & 9.91 & 19.77 & 84.03\\
\bottomrule
\end{tabular}
\caption{Response-level evaluation results on the ESC test set.}
\label{tbl:result_escov}
\end{table*}

\begin{table*}[!ht]
\centering
\small
\begin{tabular}{l|lllllllllll}
\toprule
\bf Model & \bf Avg. Score & \bf Media & \bf Min & \bf Max & \bf Success & \bf Failure & \bf S & \bf A & \bf B & \bf C & \bf F \\
\midrule
Qwen3.6-Plus & 66.4 & 80 & 0 & 100 & 13 & 14 & 13 & 49 & 16 & 8 & 14\\
$~~~~$+ESC-Skills & 72.1 & 90 & 0 & 100 & 31 & 12 & 31 & 36 & 11 & 10 & 12\\
\midrule
GPT-5.4-0305-global & 56.9 & 68 & 0 & 100 & 6 & 21 & 6 & 43 & 20 & 10 & 21\\
$~~~~$+ESC-Skills & 57.4 & 75 & 0 & 100 & 7 & 19 & 7 & 46 & 13 & 15 & 19\\
\midrule
Gemini-3.1-Flash & 56.2 & 70 & 0 & 100 & 4 & 21 & 4 & 48 & 13 & 14 & 21\\
$~~~~$+ESC-Skills & 57.6 & 75 & 0 & 100 & 7 & 19 & 7 & 42 & 19 & 13 & 19\\
\midrule
Claude-opus-4.6 & 61.20 & 75 & 0 & 100 & 16 & 18 & 16 & 40 & 13 & 13 & 18\\
$~~~~$+ESC-Skills & 61.80 & 80 & 0 & 100 & 19 & 21 & 19 & 39 & 11 & 10 & 21\\
\midrule
Claude-sonnet-4.6 & 58.20 & 70 & 0 & 100 & 9 & 23 & 9 & 44 & 15 & 9 & 23\\
$~~~~$+ESC-Skills & 63.60 & 85 & 0 & 100 & 11 & 21 & 11 & 53 & 6 & 9 & 21\\
\midrule
Claude-haiku-4.5 & 29.70 & 6 & 0 & 100 & 2 & 51 & 2 & 18 & 16 & 13 & 51\\
$~~~~$+ESC-Skills & 42.30 & 15 & 0 & 100 & 8 & 43 & 8 & 33 & 4 & 12 & 43\\
\bottomrule
\end{tabular}
\caption{Dialogue-level evaluation results on the SAGE benchmark.}
\label{tbl:result_sage}
\end{table*}

\begin{table*}[!ht]
\centering
\small
\begin{tabular}{l|lllll|lll}
\toprule
\multirow{2}{*}{\bf Model} & \multicolumn{5}{c|}{\bf ESConv} & \multicolumn{3}{c}{\bf SAGE}\\
& \bf ACC & \bf B-4 & \bf R-L & \bf METEOR & \bf BT & \bf Avg. Score & \bf Success & \bf Failure \\
\midrule
Qwen3.6-Plus & 11.50 & 0.58 & 8.92 & 14.96 & 83.73 & 66.4 & 13 & 14\\
$~~~~$+ESC-Skills($\mathcal{B}^{0}$)& 11.10 & 0.67 & 9.09 & 15.56 & 83.78 & 61.1 & 15 & 19 \\
$~~~~$+ESC-Skills($\mathcal{B}^{\triangleright}$) & 12.42 & 0.59 & 8.89 & 14.84 & 83.75 & 66.1 & 15 & 16\\
$~~~~$+ESC-Skills($\mathcal{B}^{\star}$) & 23.56 & 0.90 & 10.22 & 20.32 & 84.24 & 72.1 & 31 & 12\\
\bottomrule
\end{tabular}
\caption{Performance comparison of different skill-based baselines using Qwen3.6-Plus on the ESConv test set and the SAGE benchmark.}
\label{tbl:ablation}
\end{table*}

\section{Key Prompts}
\label{apx:prompts}
To make our pipeline reproducible, this appendix lists the verbatim prompt templates underlying the key LLM calls in ESC-Skills.


\begin{figure*}[h]
\centering
\begin{lstlisting}[
  basicstyle=\scriptsize\ttfamily,
  breaklines=true,
  columns=fullflexible,
  keepspaces=true,
  frame=single,
  framesep=4pt,
  xleftmargin=6pt,
  xrightmargin=6pt,
  aboveskip=4pt,
  belowskip=4pt,
]
# Role
You are the supporter in a two-person conversation. The seeker shares their current
problem and emotional state. Your task is to apply empathy, build emotional connection,
and provide appropriate comfort and support based on the conversation.

## Strategy Definitions
[Question]: Ask open-ended questions to explore the user's feelings and situation.
[Restatement or Paraphrasing]: Rephrase what the user said to confirm understanding
and show you are listening.
[Reflection of feelings]: Acknowledge and validate the user's emotions to show empathy.
[Self-disclosure]: Share relevant personal experiences or information when appropriate.
[Affirmation and Reassurance]: Provide comfort and reassurance to reduce the user's
anxiety or distress.
[Providing Suggestions]: Offer practical advice or suggestions to help address the
user's concerns.
[Information]: Provide factual information or explanations relevant to the situation.
[Others]: Responses that do not fit the above categories.

## OutputFormat
Choose only one strategy that aligns with the dialogue context, and craft your reply
accordingly. Strictly follow the JSON format below.
```json
{
    "strategy": "your strategy",
    "text": "your response"
}
```

{skills_section}
\end{lstlisting}
\caption{ESC Agent system prompt. The \texttt{\{skills\_section\}} placeholder is replaced by the full skill-bank content ($\mathcal{B}^{0}$, $\mathcal{B}^{\triangleright}$, or $\mathcal{B}^{\star}$) or left empty for the no-skill baseline.}
\label{fig:prompt-agent}
\end{figure*}


\begin{figure*}[h]
\centering
\begin{lstlisting}[
  basicstyle=\scriptsize\ttfamily,
  breaklines=true,
  columns=fullflexible,
  keepspaces=true,
  frame=single,
  framesep=4pt,
  xleftmargin=6pt,
  xrightmargin=6pt,
  aboveskip=4pt,
  belowskip=4pt,
]
You are an expert in Emotional Support Counseling skill design.

Below is a SAGE evaluation session for a help-seeker profile. The ESC Agent conducted
one or more conversations with this seeker. Your job is to analyze the conversations
and determine whether the current ESC skill bank needs improvement.

## Help-Seeker Profile
**Task/Hidden Theme**: {task}
**Scene Summary** (first 500 chars): {scene_summary}

## Current Skills Bank (27 skills)
{skills_catalog}

## Skills Actually Used in These Conversations
{skills_used_list}

## Full Content of Used Skills
{used_skills_content}

## Conversation Results
{conversations_text}

## Instructions

Analyze the conversations and their emotion scores. The conversations include the
evaluator's internal reasoning:
- **[Emotion Analysis]**: The scorer's analysis of how the Agent's reply affected the
  seeker's emotion, with the emotion change value.
- **[Seeker Thinking]**: The simulated seeker's internal thoughts before replying,
  revealing what they truly felt about the Agent's response.

Use these insights to understand WHY scores changed. Consider:
1. Which conversations scored well vs poorly, and why?
2. Did the agent select appropriate skills? Were the skills effective?
3. Are there gaps in the current skills bank that caused poor performance?
4. Could any existing skill be improved to handle this scenario better?

Choose ONE recommendation:
- **no_action**: The conversation went well (high score), no changes needed.
- **update_existing**: An existing skill was used but underperformed; specify which
  skill and why it needs improvement.
- **add_new**: No existing skill adequately covers this scenario; propose a new skill
  name and description.

Output a JSON object:
```json
{
    "profile_id": "<profile_id>",
    "avg_score": <average emotion score across conversations>,
    "analysis": "<2-3 sentence analysis of agent performance>",
    "skills_actually_used": ["<list of skills used>"],
    "skill_effectiveness": "<assessment of how well used skills worked>",
    "skill_gaps": ["<list of specific gaps or weaknesses found>"],
    "recommendation": "<one of: no_action | update_existing | add_new>",
    "target_skill": "<skill name if update_existing, else null>",
    "update_reason": "<why this skill needs update, if applicable>",
    "new_skill_name": "<proposed name if add_new, else null>",
    "new_skill_description": "<1-sentence description if add_new, else null>",
    "reasoning": "<detailed reasoning for your recommendation>"
}
```

IMPORTANT: Output ONLY the JSON object, no other text.
\end{lstlisting}
\caption{Skill evolution: per-profile analysis prompt. This is the first step of our evolution pipeline. The analyzer LLM digests evaluation conversations and emits a structured recommendation (\texttt{no\_action}, \texttt{update\_existing}, or \texttt{add\_new}).}
\label{fig:prompt-analysis}
\end{figure*}


\begin{figure*}[h]
\centering
\begin{lstlisting}[
  basicstyle=\scriptsize\ttfamily,
  breaklines=true,
  columns=fullflexible,
  keepspaces=true,
  frame=single,
  framesep=4pt,
  xleftmargin=6pt,
  xrightmargin=6pt,
  aboveskip=4pt,
  belowskip=4pt,
]
You are an expert Emotional Support Counseling skill designer.

You need to UPDATE an existing skill based on analysis from real evaluation sessions.

## Current SKILL.md Content
```markdown
{current_content}
```

## Why This Skill Needs Updating
{update_reason}

## Key Improvements Needed
{key_improvements}

## Evidence (from {evidence_count} evaluation profiles, avg emotion score:
{avg_score:.1f})

## Relevant Conversation Examples
{conversation_examples}

## Instructions

Rewrite the SKILL.md with the requested improvements. You must:
1. Keep the same YAML frontmatter format (name, description, metadata)
2. UPDATE the "description" field to reflect the enhanced skill capabilities
3. Increment the version number (e.g., "1.0" -> "2.0")
4. Preserve the overall structure (sections, headers)
5. Integrate the improvements naturally into the existing content
6. Keep the same domain and category
7. Add new sections/templates/examples as needed for the improvements
8. Do NOT remove existing good content -- only enhance it

Output ONLY the complete updated SKILL.md content (starting with --- frontmatter).
Do not wrap in code fences.
\end{lstlisting}
\caption{Skill evolution: update prompt. When the aggregated evolution plan calls for editing an existing skill, the generator LLM rewrites the \texttt{SKILL.md} while preserving its YAML frontmatter contract.}
\label{fig:prompt-update}
\end{figure*}


\begin{figure*}[h]
\centering
\begin{lstlisting}[
  basicstyle=\scriptsize\ttfamily,
  breaklines=true,
  columns=fullflexible,
  keepspaces=true,
  frame=single,
  framesep=4pt,
  xleftmargin=6pt,
  xrightmargin=6pt,
  aboveskip=4pt,
  belowskip=4pt,
]
You are an expert Emotional Support Counseling skill designer.

You need to CREATE a new skill for the ESC skills bank.

## Reference SKILL.md (for format guidance)
```markdown
{reference_skill}
```

## New Skill Requirements
- **Name**: {skill_name}
- **Description**: {description}
- **Rationale**: {rationale}
- **Evidence**: Requested by {evidence_count} evaluation profiles (avg emotion
  score: {avg_score:.1f})

## Relevant Conversation Examples (showing gaps the new skill should address)
{conversation_examples}

## Current Skills Bank (to avoid overlap)
{skills_list}

## Instructions

Create a complete SKILL.md for this new skill. You must:
1. Use the exact YAML frontmatter format shown in the reference (name,
   description, metadata with domain, category, version, techniques)
2. Set version to "1.0"
3. Set domain to "emotional-support-counseling"
4. Include these sections:
   - Technique Overview
   - When to Use (conditions/states)
   - Operational Steps (with language templates)
   - Contrastive Examples (good vs bad)
   - Coordination with Other Skills
5. Make language templates practical and directly usable
6. Ensure the skill fills a genuine gap not covered by existing skills

Output ONLY the complete SKILL.md content (starting with --- frontmatter).
Do not wrap in code fences.
\end{lstlisting}
\caption{Skill evolution: creation prompt. When the plan requests a brand-new skill, the generator receives a reference \texttt{SKILL.md}, a low-scoring conversation as evidence, and the existing skill catalogue to avoid overlap.}
\label{fig:prompt-create}
\end{figure*}


\begin{figure*}[h]
\centering
\begin{lstlisting}[
  basicstyle=\scriptsize\ttfamily,
  breaklines=true,
  columns=fullflexible,
  keepspaces=true,
  frame=single,
  framesep=4pt,
  xleftmargin=6pt,
  xrightmargin=6pt,
  aboveskip=4pt,
  belowskip=4pt,
]
You are an expert in emotional support counseling. Your task is to create a
comprehensive ESC skill document in SKILL.md format using a structured
chain-of-thought workflow.

Follow these five steps IN ORDER. Show your thinking for Steps 1-4 inside
<thinking>...</thinking> tags, then output the final SKILL.md content in Step 5.

## Step 1: Task Analysis
Analyze the task requirements:
- **Domain**: Emotional Support Counseling -- identify the core competencies
- **Tools**: What counseling strategies/techniques should the skill cover?
  (e.g., empathic reflection, cognitive reframing, problem-solving, validation)
- **Output Format**: SKILL.md with YAML frontmatter + markdown content
- **Pitfalls**: What are common mistakes in ESC that the skill must explicitly
  address? (e.g., premature advice-giving, emotional invalidation, projection)

## Step 2: Skill Architecture Design
Plan a single comprehensive skill covering the full ESC process:
- Define the scope: opening -> exploration -> intervention -> closing
- Identify key decision points at each stage
- Plan how strategies map to different emotional states and problem types
- Consider edge cases: crisis situations, resistance, ambivalence, silence

## Step 3: Write Skills with Progressive Disclosure
Generate the SKILL.md content with these components:
(a) **YAML frontmatter**: name and description
(b) **Key constraints and rules**: Core principles that must never be violated
(c) **Step-by-step workflow with decision points**: Stage-by-stage guidance with
    clear entry/exit criteria and branching logic
(d) **Common mistakes to avoid and edge cases**: Specific pitfalls with concrete
    examples of what NOT to do
(e) **Reusable response patterns**: Template responses for common scenarios

## Step 4: Self-Verify
Re-read the instruction and check:
- Does the skill cover ALL phases of ESC (opening through closing)?
- Are there concrete, actionable guidelines (not just abstract principles)?
- Are common pitfalls explicitly listed with examples?
- Does it handle edge cases (crisis, resistance, silence)?
- Is the YAML frontmatter valid?
- Is the length between 200-400 lines?

## Step 5: Execute
Output ONLY the final SKILL.md content, starting with `---` (the YAML front matter
delimiter).
\end{lstlisting}
\caption{Self-generated skill baseline: CoT-guided mode. The model is asked to produce a comprehensive ESC skill from scratch using a five-step chain-of-thought workflow. This serves as the \texttt{cot} baseline in our ablation study.}
\label{fig:prompt-selfgen-cot}
\end{figure*}


\begin{figure*}[h]
\centering
\begin{lstlisting}[
  basicstyle=\scriptsize\ttfamily,
  breaklines=true,
  columns=fullflexible,
  keepspaces=true,
  frame=single,
  framesep=4pt,
  xleftmargin=6pt,
  xrightmargin=6pt,
  aboveskip=4pt,
  belowskip=4pt,
]
You are an expert evaluator for Emotional Support Conversations (ESC). Your task is
to evaluate the quality of a supporter's response in a counseling-style dialogue.

You will be given:
1. The seeker's situation and emotional context
2. The conversation history
3. The supporter's response to evaluate

Rate the response on four dimensions using a 1-5 Likert scale:

## Scoring Rubric

### Empathy
- 5: Deeply attuned to emotions; validates feelings with genuine warmth; makes
     seeker feel truly heard
- 4: Shows clear emotional understanding; acknowledges feelings appropriately
- 3: Basic emotional acknowledgment; somewhat formulaic but acceptable
- 2: Superficial or generic; misses emotional nuances
- 1: Cold, dismissive, or emotionally tone-deaf

### Relevance
- 5: Perfectly addresses the current conversational need; builds naturally on context
- 4: Clearly relevant; responds to the main point with appropriate depth
- 3: Generally on-topic but may miss some contextual details
- 2: Partially off-topic or too generic for the specific situation
- 1: Irrelevant, ignores context, or introduces confusing tangents

### Helpfulness
- 5: Provides meaningful support that could genuinely help; advances the conversation
     productively
- 4: Offers useful support; helps seeker explore or cope with their situation
- 3: Mildly helpful; provides some support but limited depth or actionability
- 2: Minimally helpful; too vague or prescriptive without understanding
- 1: Unhelpful or potentially harmful; shuts down conversation or gives inappropriate
     advice

### Overall
- 5: Excellent emotional support response -- empathetic, relevant, and genuinely
     helpful
- 4: Good response that serves the seeker's needs well
- 3: Adequate response; does the job but nothing remarkable
- 2: Below average; has notable weaknesses
- 1: Poor response; fails as emotional support

## Output Format
You MUST respond in the following JSON format only:
```json
{
    "empathy": <int 1-5>,
    "relevance": <int 1-5>,
    "helpfulness": <int 1-5>,
    "overall": <int 1-5>,
    "rationale": "<brief 1-2 sentence justification>"
}
```

--- User Message Template ---

## Seeker's Situation
{situation}

## Conversation History
{history}

## Supporter's Response to Evaluate
Strategy: [{strategy}]
Response: {response}

Please evaluate this response.
\end{lstlisting}
\caption{LLM-as-Judge evaluation prompt (system + user template). GPT-5.4 scores each candidate response on \textit{empathy}, \textit{relevance}, \textit{helpfulness} and \textit{overall} using a 1--5 Likert scale.}
\label{fig:prompt-judge}
\end{figure*}

\section{Human Evaluation Annotation Guidelines}
\label{apx:annotation}

Figure~\ref{fig:annotation-guideline} presents the annotation instructions shown to human evaluators. The guidelines cover the task setup, rating dimensions, annotator qualifications, quality-control procedures, and ethical considerations. Annotators rated each response on empathy, helpfulness, and overall quality using a 1–5 Likert scale. Model identities were hidden, and response order was randomized to reduce annotation bias.


\begin{figure*}[h]
\centering
\begin{lstlisting}[
  basicstyle=\scriptsize\ttfamily,
  breaklines=true,
  columns=fullflexible,
  keepspaces=true,
  frame=single,
  framesep=4pt,
  xleftmargin=6pt,
  xrightmargin=6pt,
  aboveskip=4pt,
  belowskip=4pt,
]
# Human Evaluation Annotation Guidelines

## Task Description
You will rate supporter responses generated by different ESC models on the ESConv
test set. For each item, you are presented with:
1. The seeker's situation description
2. The dialogue history up to the current turn
3. A single supporter response to evaluate

Model identities are HIDDEN, and responses from different models for the same
context are randomized across the annotation queue to mitigate ordering and
identification bias.

## Rating Dimensions
Rate each response on three dimensions using a 1-5 Likert scale.

### Empathy
- 5: Deeply attuned; validates feelings with genuine warmth; makes the seeker
     feel truly heard
- 4: Shows clear emotional understanding; acknowledges feelings appropriately
- 3: Basic emotional acknowledgment; somewhat formulaic but acceptable
- 2: Superficial or generic; misses emotional nuances
- 1: Cold, dismissive, or emotionally tone-deaf

### Helpfulness
- 5: Provides meaningful support that could genuinely help; advances the
     conversation productively
- 4: Offers useful support; helps seeker explore or cope with their situation
- 3: Mildly helpful; limited depth or actionability
- 2: Minimally helpful; too vague or prescriptive without understanding
- 1: Unhelpful or potentially harmful

### Overall Quality
A holistic 1-5 rating capturing the response's value as emotional support,
integrating empathy, contextual relevance, and helpfulness.

## Annotator Background
Three independent annotators participated in the evaluation. All hold at least
a bachelor's degree, are proficient in English, and received a 30-minute
training session with five calibration examples before the formal annotation.
Annotators were compensated above the local minimum wage.

## Quality Control
- Calibration round: Five practice items with reference scores were used to
  align annotators before the main task.
- Attention checks: 5% of items were duplicated; annotators with intra-rater
  divergence >= 2 on more than 20% of duplicates were excluded.
- Aggregation: Final scores per item are the mean of three annotators;
  inter-annotator agreement is reported as Fleiss' kappa on the discretized
  ratings.

## Ethical Considerations
The ESConv corpus contains discussions of emotionally sensitive topics.
Annotators were informed of this in advance, given the option to opt out at any
time, and provided with mental-health resource references. No personally
identifying information was shown.
\end{lstlisting}
\caption{Human evaluation annotation guidelines. Three independent annotators rated the same 100 sampled ESConv supporter turns used for GPT-Judge scoring, covering three representative model pairs on \textit{empathy}, \textit{helpfulness} and \textit{overall quality} using a 1--5 Likert scale.}
\label{fig:annotation-guideline}
\end{figure*}

\end{document}